%% file: main.tex
\documentclass[pdflatex,sn-nature]{sn-jnl}

\usepackage{graphicx}%
\usepackage{multirow}%
\usepackage{amsmath,amssymb,amsfonts}%
\usepackage{amsthm}%
\usepackage{mathrsfs}%
\usepackage[title]{appendix}%
\usepackage{xcolor}%
\usepackage{textcomp}%
\usepackage{manyfoot}%
\usepackage{booktabs}%
\usepackage{algorithm}%
\usepackage{algorithmicx}%
\usepackage{algpseudocode}%
\usepackage{listings}%
\input{content/x_setup}

\theoremstyle{thmstyleone}%
\theoremstyle{thmstyletwo}%

\theoremstyle{thmstylethree}%

\begin{document}

\title{Learning to Forget -- Hierarchical Episodic Memory for Lifelong Robot Deployment
}

\author*{\fnm{Leonard} \sur{Bärmann}}\email{baermann@kit.edu}
\author{\fnm{Joana} \sur{Plewnia}}
\author{\fnm{Alex} \sur{Waibel}}
\author*{\fnm{Tamim} \sur{Asfour}}\email{asfour@kit.edu}

\affil{\orgdiv{Institute for Anthropomatics and Robotics}, \orgname{Karlsruhe Institute for Technology}, \orgaddress{
\country{Germany}}}

\abstract{
\input{content/0_abstract}
}


\maketitle

\input{figures/teach-eval.tikz}
\input{figures/dataset-sample.tikz}

\input{content/1_content}

\section*{Acknowledgments}
This work has been supported by the Baden-Württemberg Ministry of Science, Research and the Arts (MWK) as part of the state's ``digital@bw'' digitization strategy in the context of the Real-World Lab ``Robotics AI'', the Carl Zeiss Foundation through the JuBot project and the German Federal Ministry of Research, Technology and Space (BMFTR) under the Robotics Institute Germany (RIG).

\clearpage
\begin{appendices}
\crefalias{section}{appendix}
\crefalias{subsection}{appendix}

\input{content/2_appendix}
\end{appendices}

\clearpage
\bibliography{main}

\end{document}

%% file: content/x_setup.tex
\usepackage{subcaption}
\usepackage{amsmath}
\usepackage{amssymb}
\usepackage{hyperref}
\usepackage{cleveref}
\usepackage{xspace}
\usepackage{xparse}
\usepackage{adjustbox}
\usepackage{mdframed}
\usepackage{tcolorbox}
\usepackage{pgfplots}
\pgfplotsset{compat=1.18}
\tcbuselibrary{skins, breakable}
\usetikzlibrary{fit,shapes.callouts, arrows.meta, positioning,calc,backgrounds}

\usepackage{pifont}
\newcommand{\cmark}{\ding{51}}
\newcommand{\xmark}{\ding{55}}

\usepackage{algorithm}
\usepackage{algpseudocode}

\newcommand{\algorithmicbreak}{\textbf{break}}
\newcommand{\Break}{\State \algorithmicbreak}
\newcommand{\algsize}[0]{\fontsize{9}{9.5} \selectfont}

\usepackage[
    frozencache=true,  
    cachedir=minted-cache,
]{minted}
\usepackage{csquotes}

\newcommand{\ie}{i.\,e.,\xspace}
\newcommand{\eg}{e.\,g.,\xspace}

\newcommand{\armarVII}{\mbox{\textsc{Armar}-7}\xspace}

\newcommand{\ours}{\textsc{H\textsuperscript{2}-Emv}\xspace}
\newcommand{\oursInHeader}{H\textsuperscript{2}-EMV\xspace}

\definecolor{mediumgray}{gray}{0.6}
\newcommand{\noforget}[1]{\textcolor{mediumgray}{#1}}

\newcommand{\variantName}[1]{\textit{#1}\xspace}
\newcommand{\varA}[0]{\variantName{ours}}
\newcommand{\varAA}[0]{\variantName{no summ. learn.}}
\newcommand{\varB}[0]{\variantName{no learning}}
\newcommand{\varC}[0]{\variantName{learn only summ.}}
\newcommand{\varD}[0]{\variantName{time-only forg.}}
\newcommand{\varE}[0]{\variantName{\noforget{no f. learn summ.}}}
\newcommand{\varF}[0]{\variantName{\noforget{no forg.}}}
\newcommand{\varG}[0]{\variantName{offl. full}}
\newcommand{\varH}[0]{\variantName{offl. no learning}}
\newcommand{\varI}[0]{\variantName{offl. time-only forg.}}
\newcommand{\varJ}[0]{\variantName{\noforget{offl. no forg.}}}
\newcommand{\varK}[0]{\variantName{flat full}}
\newcommand{\varL}[0]{\variantName{\noforget{flat no forg.}}}

\newcommand{\twocol}[1]{\multicolumn{2}{c}{#1}}
\newcommand{\twocolb}[1]{\multicolumn{2}{c|}{#1}}

\crefname{section}{Section}{Sections}
\Crefname{section}{Section}{Sections}
\crefname{appendix}{Appendix}{Appendices}
\Crefname{appendix}{Appendix}{Appendices}
\crefname{figure}{Figure}{Figures}
\Crefname{figure}{Figure}{Figures}
\crefname{table}{Table}{Table}
\Crefname{table}{Table}{Table}
\crefname{equation}{Eq.}{Eqs.}
\Crefname{equation}{Eq.}{Eqs.}
\crefname{listing}{Listing}{Listings}
\Crefname{listing}{Listing}{Listings}
\crefname{algorithm}{Algorithm}{Algorithms}
\Crefname{algorithm}{Algorithm}{Algorithms}

\makeatletter
\newenvironment{chatbox}[2][text]{
    \VerbatimEnvironment
    \footnotesize
    \begin{mdframed}[
        frametitle={
            \tcbox[on line, boxsep=4pt, left=0pt,right=0pt,top=0pt,bottom=0pt,colframe=white,colback=gray]{
                \textcolor{white}{\space#2\space}
            }
        },
        innertopmargin=-0pt,
        frametitleaboveskip=-3ex,
    ]
    \begin{minted}[breaklines]{#1}%
}{%
    \end{minted}%
    \end{mdframed}\filbreak
    \vspace{-1.5em}
}
\makeatother

%% file: content/0_abstract.tex
Robots must verbalize their past experiences when users ask \enquote{Where did you put my keys?} or \enquote{Why did the task fail?} 
Yet maintaining life-long episodic memory~(EM) from continuous multimodal perception quickly exceeds storage limits and makes real-time query impractical, calling for selective forgetting that adapts to users’ notions of relevance. 
We present \ours, a framework enabling humanoids to learn what to remember through user interaction. 
Our approach incrementally constructs hierarchical EM, selectively forgets using language-model-based relevance estimation conditioned on learned natural-language rules, and updates these rules given user feedback about forgotten details. 
Evaluations on simulated household tasks and 20.5-hour-long real‑world recordings from \armarVII demonstrate that \ours maintains question-answering accuracy while reducing memory size by 45\% and query-time compute by 35\%. 
Critically, performance improves over time---accuracy increases 70\% in second-round queries by adapting to user-specific priorities---demonstrating that learned forgetting enables scalable, personalized EM for long-term human-robot collaboration.

%% file: figures/teach-eval.tikz
\newcommand{\plotbarwidth}[0]{3.5pt}
\newcommand{\plotbarshift}[0]{1.75pt}
\newcommand{\enlargeylimitsamount}[0]{0.08}
\definecolor{plotgreenbefore}{HTML}{61D836}
\definecolor{plotgreenafter}{HTML}{5D9648}
\definecolor{plotorangebefore}{HTML}{F8BA00}
\definecolor{plotorangeafter}{HTML}{E7A13D}
\definecolor{plotblueone}{HTML}{56C1FF}
\definecolor{plotbluetwo}{HTML}{0076BA}
\definecolor{plotother}{HTML}{00AB7A}

\NewDocumentCommand{\implBeforeAfterStackedBarPlot}{m m m m +m +m O{1.7} O{} O{1}}{%
  \DeclareDocumentCommand{\implimplBeforeAfterStackedBarPlot}{O{from table={#3}{Label}}}{%
  \begin{tikzpicture}[
    baseline=(current bounding box.south),
  ]
    \begin{axis}[
      name=main,
      xbar,
      xbar stacked,
      width=#9\linewidth,
      height=#7,
      xmin=0, xmax={#1},
      axis x line=bottom,
      axis y line=left,
      ytick=data,                              
      yticklabels ##1,      
      yticklabel style={font=\small},
      ytick style={draw=none},
      y axis line style={-}, 
      y dir=reverse,
      enlarge y limits=\enlargeylimitsamount,
      xlabel={#4},
      bar width=\plotbarwidth,
      legend style={
        at={(0.5,1)}, anchor=south, legend columns=2, draw=none
      },
      legend image code/.code={
        \draw[draw=none] (0cm,-0.3em) rectangle (1em,0.7em);
      },
      grid=major,
      xmajorgrids=true,
      grid style={gray!30},
      table/col sep=tab
    ]

      #5

      \draw[thick] (rel axis cs:0,#2) -- (rel axis cs:1,#2);

      \if\relax\detokenize{#8}\relax
      \else
        \draw[thick] (rel axis cs:0,#8) -- (rel axis cs:1,#8); %
      \fi
    
    \end{axis}

    \begin{axis}[
      at={(main.south west)}, anchor=south west,
      xbar,
      xbar stacked,
      width=#9\linewidth, 
      height=#7,
      xmin=0, xmax=#1,
      axis lines=none,
      ytick=data,
      y dir=reverse,
      enlarge y limits=\enlargeylimitsamount,
      bar width=\plotbarwidth,
      table/col sep=tab
    ]
      #6
    \end{axis}
  \end{tikzpicture}%
  }
  \implimplBeforeAfterStackedBarPlot
}

\NewDocumentCommand{\QaPerformanceGroupedStackedBarPlot}{O{80} O{0.4} m O{1} O{} O{1} O{}}{%
  \implBeforeAfterStackedBarPlot{#1}{#2}{#3}{Performance [\%]}{
      \addplot+[fill=plotgreenbefore, draw=none, bar shift=+\plotbarshift]
        table[x={QA S_c before}, y expr=\coordindex, col sep=tab] {#3};
      
      \if\relax\detokenize{#7}\relax
        \addlegendentry{corr. round 1}
      \else
      \fi

      \addplot+[fill=plotorangebefore, draw=none, bar shift=+\plotbarshift]
        table[x={S_p' before}, y expr=\coordindex, col sep=tab] {#3};
      \if\relax\detokenize{#7}\relax
        \addlegendentry{part. corr. round 1}
      \else
      \fi

      \if\relax\detokenize{#7}\relax
        \addlegendimage{draw=none, fill=plotgreenafter}
        \addlegendentry{corr. round 2}
        \addlegendimage{draw=none, fill=plotorangeafter}
        \addlegendentry{part. corr. round 2}
      \else
      \fi

  }{
      \addplot+[fill=plotgreenafter, draw=none, bar shift=-\plotbarshift]
        table[x={QA S_c after},  y expr=\coordindex, col sep=tab] {#3};
      \addplot+[fill=plotorangeafter, draw=none, bar shift=-\plotbarshift]
        table[x={S_p' after},    y expr=\coordindex, col sep=tab] {#3};
  }[#4][#5][#6]
}

\NewDocumentCommand{\TreeSizeGroupedBarPlot}{O{0.4} m m O{1} O{} O{1} O{}}{%
  \implBeforeAfterStackedBarPlot{#2}{#1}{#3}{Tree size [\# L3+ nodes]}{
      \addplot+[fill=plotblueone, draw=none, bar shift=+\plotbarshift]
        table[x={avg tree size}, y expr=\coordindex, col sep=tab] {#3};
      
      \if\relax\detokenize{#7}\relax
        \addlegendentry{avg. tree}
        \addlegendimage{draw=none, fill=plotbluetwo}
        \addlegendentry{final tree}
      \else
      \fi
  }{
      \addplot+[fill=plotbluetwo, draw=none, bar shift=-\plotbarshift]
        table[x={final tree size},  y expr=\coordindex, col sep=tab] {#3};
  }[#4][#5][#6][={}]
}

\NewDocumentCommand{\FinalTreeSizeSimpleBarPlot}{O{0.4} m m O{1.7} O{} O{1} O{}}{%
  \begin{tikzpicture}[baseline=(current bounding box.south)]
    \begin{axis}[
      xbar,
      width=#6\linewidth,
      height=#4,
      xmin=0, xmax={#2},
      axis x line=bottom,
      axis y line=left,
      ytick=data,                              
      yticklabels={},
      ytick style={draw=none},
      y axis line style={-}, 
      y dir=reverse,
      enlarge y limits=\enlargeylimitsamount,
      xlabel={Tree size [\# L3+ nodes]},
      bar width=\plotbarwidth,
      legend style={
        at={(0.5,1)}, anchor=south, legend columns=2, draw=none
      },
      legend image code/.code={
        \draw[fill=##1,draw=none] (0cm,-0.1cm) rectangle (0.3cm,0.2cm);
      },
      grid=major,
      xmajorgrids=true,
      grid style={gray!30},
      table/col sep=tab
    ]

      \addplot+[fill=plotbluetwo, draw=none]
        table[x={final tree size},  y expr=\coordindex, col sep=tab] {#3};
      \if\relax\detokenize{#7}\relax
        \addlegendentry{final tree size}
      \else
      \fi

      \draw[thick] (rel axis cs:0,#1) -- (rel axis cs:1,#1);

      \if\relax\detokenize{#5}\relax
      \else
        \draw[thick] (rel axis cs:0,#5) -- (rel axis cs:1,#5); %
      \fi

    \end{axis}
  \end{tikzpicture}
}

\NewDocumentCommand{\QaTokensBarPlot}{O{0.4} m m O{1.7} O{} O{1} O{}}{%
  \begin{tikzpicture}[baseline=(current bounding box.south)]
    \begin{axis}[
      xbar,
      width=#6\linewidth,
      height=#4,
      xmin=0, xmax={#2},
      axis x line=bottom,
      axis y line=left,
      ytick=data,                              
      yticklabels={},
      ytick style={draw=none},
      y axis line style={-}, 
      y dir=reverse,
      enlarge y limits=\enlargeylimitsamount,
      xlabel={QA tokens [1K]},
      bar width=\plotbarwidth,
      legend style={
        at={(0.5,1)}, anchor=south, legend columns=2, draw=none
      },
      legend image code/.code={
        \draw[fill=##1,draw=none] (0cm,-0.1cm) rectangle (0.3cm,0.2cm);
      },
      grid=major,
      xmajorgrids=true,
      grid style={gray!30},
      table/col sep=tab
    ]

      \addplot+[fill=plotother, draw=none]
        table[x={QA tokens}, y expr=\coordindex, col sep=tab] {#3};
      \if\relax\detokenize{#7}\relax
        \addlegendentry{QA tokens}
      \else
      \fi

      \draw[thick] (rel axis cs:0,#1) -- (rel axis cs:1,#1);

      \if\relax\detokenize{#5}\relax
      \else
        \draw[thick] (rel axis cs:0,#5) -- (rel axis cs:1,#5); %
      \fi

    \end{axis}
  \end{tikzpicture}
}

\NewDocumentCommand{\EvalOverviewPlots}{O{0.4} O{} m m m m O{1} O{}}{
    \footnotesize
    \setlength{\tabcolsep}{0pt}
    \begin{tabular}{p{0.5\linewidth} p{0.3\linewidth} p{0.2\linewidth}}
        \QaPerformanceGroupedStackedBarPlot[#3][#1]{#6}[#7][#2][0.85][#8]
        &
        \FinalTreeSizeSimpleBarPlot[#1]{#4}{#6}[#7][#2][1.2][#8]
        &
        \QaTokensBarPlot[#1]{#5}{#6}[#7][#2][1.4][#8]
        \\
    \end{tabular}
}

\NewDocumentCommand{\EvalDetailPlots}{O{0.4} O{} m m m m O{1} O{}}{
    \footnotesize
    \setlength{\tabcolsep}{0pt}
    \begin{tabular}{p{0.57\linewidth} p{0.25\linewidth} p{0.18\linewidth}}
        \QaPerformanceGroupedStackedBarPlot[#3][#1]{#6}[#7][#2][0.8][#8]
        &
        \TreeSizeGroupedBarPlot[#1]{#4}{#6}[#7][#2][1.4][#8]
        &
        \QaTokensBarPlot[#1]{#5}{#6}[#7][#2][1.3][#8]
        \\
    \end{tabular}
}

%% file: figures/dataset-sample.tikz
\tikzset{
    qbubble/.style={
        rectangle callout,
        callout relative pointer={(-0.3,0)}, 
        callout pointer arc=0,
        fill=gray!5,
        draw=gray!80,
        rounded corners,
        align=left,
        text width=0.7\linewidth,
        anchor=west,
        inner ysep=5pt
    },
    abubble/.style={
        rectangle callout,
        callout relative pointer={(0.3,0)}, 
        callout pointer arc=0,
        fill=gray!5,
        draw=gray!80,
        rounded corners,
        align=left,
        text width=0.7\linewidth,
        anchor=west,
        inner ysep=5pt
    },
    fbubble/.style={
        rectangle callout,
        callout relative pointer={(-0.3,0)}, 
        callout pointer arc=0,
        fill=gray!5,
        draw=gray!80,
        rounded corners,
        align=left,
        text width=0.7\linewidth,
        anchor=west,
        inner ysep=5pt
    }
}

\NewDocumentCommand{\TikzTwoRoundEmvExample}{m m m m m m m m m}{%
\begin{tikzpicture}[>=Stealth, node distance=0mm and 0mm]

    \def\timelineX{-2mm}

    \node[qbubble] (q1) {$Q_1$: #2};
    \node[abubble, below=1mm of q1] (a1) {$A_1$: #4};
    \node[fbubble, below=1mm of a1] (f1) {$F$: #5};

    \node[left=4mm of q1.west] (t1) {$t_1$};
    \node[left=0mm of t1.west,anchor=center,xshift=\timelineX] (bar1) {---};
    \node[below=0mm of t1.south,anchor=west,rotate=-90] (date1) {\tiny\!=\!#1};

    \node[below=1mm of f1] (dots) {\tiny(many actions later)};

    \node[qbubble, below=6mm of f1] (q2) {$Q_2$: #7};
    \node[abubble, below=1mm of q2] (a2) {$A_2$: #9};

    \node[left=4mm of q2.west] (t2) {$t_2$};
    \node[left=0mm of t2.west,anchor=center,xshift=\timelineX] (bar2) {---};
    \node[below=0mm of t2.south,anchor=west,rotate=-90] (date2) {\tiny\!=\!#6};

    \node[draw=none, inner sep=0pt, fit=(t1)(q1)(a2)(date2)] (allBubbles) {};
    \draw[thick, ->] ([xshift=\timelineX]allBubbles.north west) -- ([xshift=\timelineX]allBubbles.south west);
                    
\end{tikzpicture}
}

\NewDocumentCommand{\TwoRoundEmvExample}{O{Example}}{%
    \DeclareDocumentCommand{\implTwoRoundEmvExample}{m m m m m m m m m}{%
        \begin{figure}[!ht]
        \centering
        \scriptsize
            \TikzTwoRoundEmvExample{##1}{##2}{##3}{##4}{##5}{##6}{##7}{##8}{##9}
        \caption{#1}
        \end{figure}
    }%
    \implTwoRoundEmvExample
}%

%% file: content/1_content.tex
\newcommand{\term}[1]{#1}

\section{Introduction}

Episodic Memory (EM) enables humans to remember, organize, and recall past events and experiences, and therefore is a key aspect in human cognition \cite{tulving_episodic_1972}.
In everyday interactions with others, we use EM to talk about and reflect upon previous experiences.
To transfer this capability to artificial agents such as humanoid robots, they need the ability to store and organize their continuous stream of incoming events in an EM and verbalize it appropriately when receiving a user's request, i.e., perform the task of EM Verbalization (EMV) \cite{rosenthal_verbalization_2016,dechant_toward_2021,barmann_deep_2021}.
EM in humans is not unlimited -- we manage information by selectively forgetting irrelevant details.
While an artificial system could outperform a human's memory here, an ever-growing EM would similarly bring any realistic system to its storage and compute limits when running for extended periods of time.
Therefore, we argue that selective forgetting is a desirable property of an artificial EM to limit storage requirements, compute at question time, and even improve precision by storing only the relevant details.
However, there is no generally accepted definition of relevance, as this may greatly differ between use cases and users of a system.
Because of this, an EM system needs to incrementally learn what its user considers relevant based on its usage and the provided feedback.

Existing systems for EMV of a robot's experiences are either end-to-end trained neural networks \cite{barmann_deep_2021,dechant_learning_2023}, or utilize text-based representations of EM in combination with natural language templates \cite{rosenthal_verbalization_2016,zhu_autonomous_2017,katuwandeniya_what_2025,plewnia_combining_2025} or LLMs \cite{liu_reflect_2023,wang_i_2024,barmann_episodic_2025} for verbalization.
To tackle the problem of scaling to very long sequences of events, previous works explored retrieval-augmented generation \cite{liu_reflect_2023} or tree-based representations of EM \cite{barmann_episodic_2025}.
While these approaches effectively structure the search problem at question time, they still produce an ever-growing EM when deployed in a lifelong setting, and will eventually also suffer from too many memories.
Further, none of these systems prioritizes memories based on an incrementally learned, user-defined relevance.

Therefore, we introduce \ours, an architecture modeling the EM of an autonomous agent, featuring 
(1) \term{online memory construction}, 
(2) \term{decay-based forgetting with relevance estimation}, and
(3) \term{incremental learning of a relevance measure from user feedback}.
Specifically, we extend the hierarchical, tree-based EM representation of \textsc{H-Emv} \cite{barmann_episodic_2025}:
First, we transfer the offline, \enquote{retrospective} EM construction of \textsc{H-Emv} to an \term{online setting by updating the history tree incrementally} as new events occur.
Secondly, inspired by the literature on forgetting \cite{plewnia_forgetting_2024, freedman_2011}, we introduce a \term{decay-based forgetting mechanism with relevance estimation}.
 Each node within the history tree is assigned a default lifetime, and the relevance for retention of expired nodes is estimated using an LLM conditioned on a set of natural-language relevance rules.
Since there is no universal set of such rules, we finally propose to \term{incrementally learn a relevance measure from user feedback}.
Specifically, when a user asks a question and the queried information has already been forgotten, they can provide feedback to improve future forgetting and summarization processes.

To evaluate our system, we use simulated recordings of household tasks from TEACh \cite{padmakumar_teach_2022} as well as real-world recordings from our humanoid robot \armarVII~\cite{asfour_armar_2017}.
Through a systematic ablation study, we show that each contribution is effective:
\term{Online history tree construction} shifts the EM construction effort away from question time, thereby reducing QA latency, while also introducing a useful recency bias.
\term{Relevance-based forgetting} effectively reduces memory size while retaining information relevant for later QA.
Finally, \term{feedback-based relevance learning} leads to notable performance improvements over time, successfully adapting to the user's needs from single or only a few feedback instances.
We further demonstrate our system deployed on \armarVII, showcasing its real-world usage while also highlighting remaining challenges for future work.


\section{Results}
\newcommand{\resultsSec}[1]{\subsubsection*{#1}}

We first describe \ours, then present simulated and real‑world experiments, and finally discuss a multi‑week deployment on the real-world humanoid robot \armarVII.

\resultsSec{Modeling \& verbalizing EM with \oursInHeader}
\begin{figure*}
    \centering
    \begin{subfigure}{1\textwidth}
        \centering
        \includegraphics[width=1\linewidth]{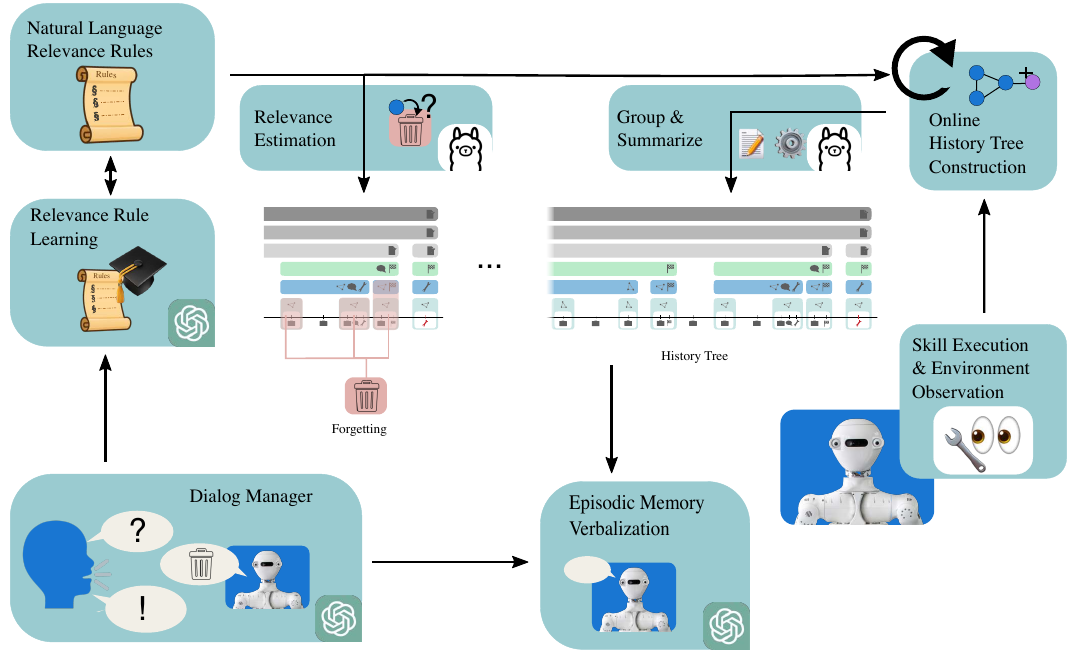}
        \caption{System architecture and typical information flow of \ours:
        From the robot's observations, \term{online history tree construction} \term{incrementally groups and summarizes} new nodes in context of existing ones.
        Nodes that reached their expiration date are passed to relevance estimation and may be forgotten.
        Both \term{online history tree construction} and relevance estimation are conditioned on a set of natural language relevance rules learned during dialog:
        Given a question (\enquote{?}), the dialog manager invokes EMV to explore the history tree.
        If relevant details are forgotten, the dialog manager indicates this (\raisebox{-2pt}{\includegraphics[height=0.95em]{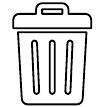}}\:\!) and the user provides feedback (\enquote{!}), triggering relevance rule learning, thereby improving future summarization and forgetting.
        }
        \label{fig:system:architecture}
    \end{subfigure}
    \\[0.3cm]
    \begin{subfigure}{0.56\textwidth}
        \centering
        \includegraphics[width=1\linewidth]{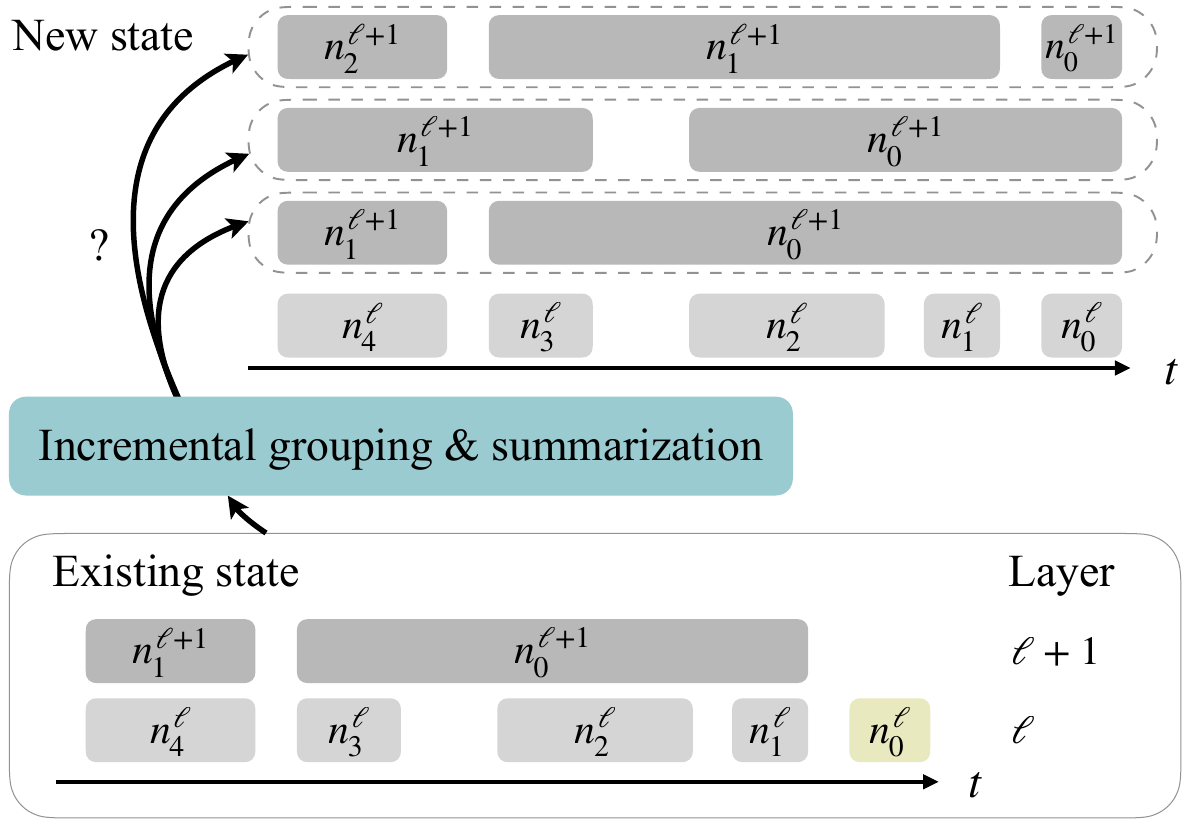}
        \caption{The \term{time-aware incremental grouping and summarization} algorithm (Alg. \ref{alg:orga:methods:time_merge}) adjusts the summaries on layer~$\ell + 1$ given a new node~$n_0^\ell$ on layer~$\ell$}
        \label{fig:system:inc_tree}
    \end{subfigure}
    \hfill
    \begin{subfigure}{0.38\textwidth}
        \centering
        \includegraphics[width=1\linewidth]{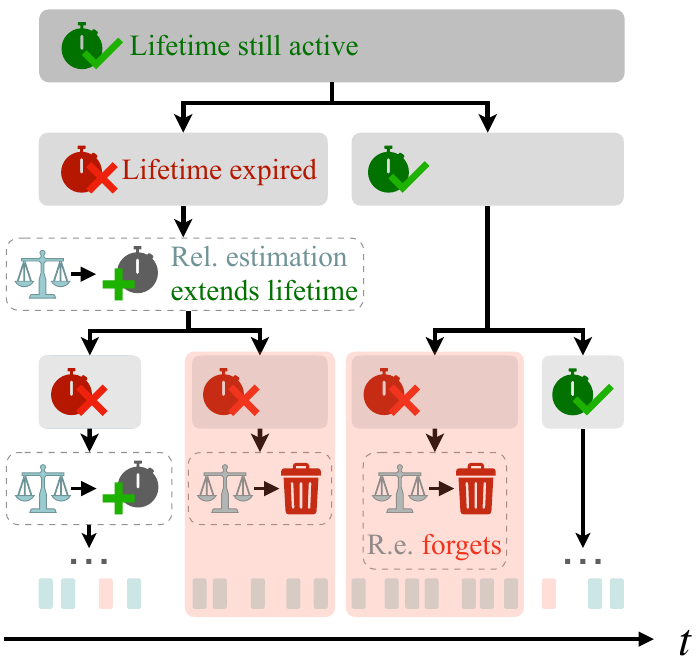}
        \caption{Forgetting inspects the history tree top-down and estimates relevance of expired nodes to extend relevant and drop irrelevant ones}
        \label{fig:system:forgetting}
    \end{subfigure}
    \caption{
    \ours models and verbalizes episodic memory of a robot using a tree-like data structure \term{constructed online}, managing storage and compute efficiency by forgetting of old items based on a \term{relevance measure learned from user feedback}.
    }
    \label{fig:system}
    
\end{figure*}
\ours equips an autonomous agent with the capability to build an EM \term{in an online setting}, prioritize relevant memories during summarization, forget irrelevant memories to save storage, efficiently answer questions about its past (EMV), and \term{incrementally learn an evolving relevance measure from task and user feedback}.
Building on previous work \cite{barmann_episodic_2025}, from a stream of observations, we derive a \emph{history tree}, a hierarchical data structure to represent EM, by recursive grouping and summarization.
Given a subsequent question, an agentic LLM is used to efficiently explore the history tree to find the answer.
\Cref{fig:system} provides an overview of our system (details in \cref{sec:methods}).

\begin{figure*}[!ht]
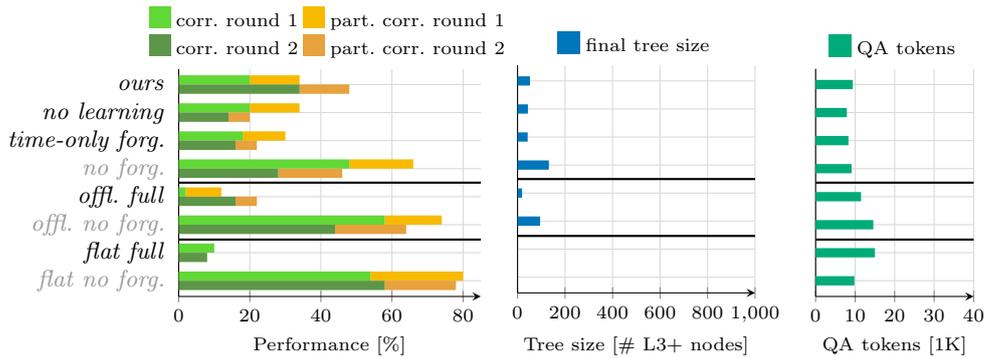
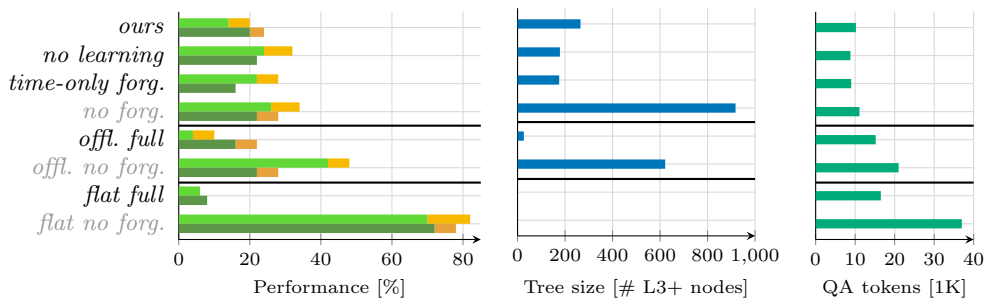
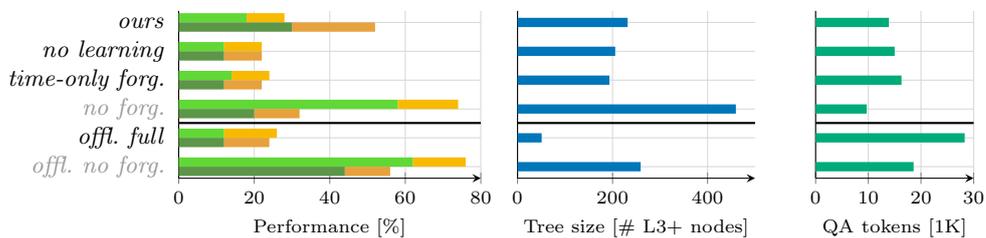

    \centering
    \begin{subfigure}{1\textwidth}
        \centering
        \scriptsize
        \setlength{\tabcolsep}{3mm}
        
        \begin{tabular}{r|c|cc|c}
                        &   history tree     &  \twocolb{forgetting based on...} & relevance \\
                        &   construction     &   ...time & ...relevance          &  learning \\\hline
             \varA      &       online       & \cmark &    \cmark    &  \cmark       \\
             \varB      &       online       & \cmark &    \cmark    &  \xmark      \\
             \varD      &       online       & \cmark &    \xmark    &  \xmark       \\
             \varF      &       online       & \xmark &              &  \xmark       \\\hline 
             \varG      &      offline       & \cmark &    \cmark    &  \cmark       \\
             \varJ      &      offline       & \xmark &              &  \xmark       \\\hline
             \varK      &   flat (L3)        & \cmark &    \cmark    &  \cmark       \\
             \varL      &   flat (L3)        & \xmark &              &  \xmark       \\
        \end{tabular}
        \caption{We compare \ours (\varA) against multiple ablations and baselines:
        \term{History tree construction} can be performed online, offline, or not at all (flat history).
        Forgetting is based on time and relevance estimation, time only, or \noforget{ablated completely}.
        Lastly, \term{feedback-based relevance learning} can be toggled.
        Further ablations, including a more detailed view on relevance learning for forgetting and summarization, can be found in \cref{app:results}.
        }
        \label{tab:orga:eval:baselines}
    
    \end{subfigure}
    \\[0.1cm]
    \begin{subfigure}{1\textwidth}
        \centering
        \EvalOverviewPlots[0.5][0.25]{85}{1000}{40}{figures/data-teach-5-simpl.tsv}[4.6cm]
        \vspace{-0.3cm}
        \caption{Results on simulated robot recordings from TEACh, histories made of $|h| = 5$ episodes}
        \label{fig:results:qa_teach_5}
    \end{subfigure}
    \\[0.3cm]
    \begin{subfigure}{1\textwidth}
        \centering
        \EvalOverviewPlots[0.5][0.25]{85}{1000}{40}{figures/data-teach-25-simpl.tsv}[4.6cm][nolegend]
        \vspace{-0.2cm}
        \caption{Results on simulated robot recordings from TEACh, histories made of $|h| = 25$ episodes}
        \label{fig:results:qa_teach_25}
    \end{subfigure}
    \\[0.3cm]
    \begin{subfigure}{1\textwidth}
        \centering
        \EvalOverviewPlots[0.33]{80}{500}{30}{figures/data-armar7-simpl.tsv}[3.8cm][nolegend]
        \vspace{-0.5cm}
        \caption{Results of experiments on real-world robot recordings from \armarVII}
        \label{fig:results:armarx}
    \end{subfigure}
    \caption{Overview of results from simulated and real-world robot experiments}
    \label{fig:results}
\end{figure*}

To quantitatively analyze the performance of our system, we evaluate it under two settings:
Using simulated agent recordings from TEACh \cite{padmakumar_teach_2022}, we randomly concatenate episodes to form long histories and automatically create diverse QA samples asking for past observations and actions.
Similarly, we create a dataset of real-world recordings from the humanoid robot \armarVII, totaling 20.5 hours of multimodal data from the robot's proprioception and perception, annotated with human-curated QA samples.
\Cref{fig:eval:procedure} illustrates the two-round evaluation procedure (details in \cref{sec:experiments}). 
We measure the evolution of QA performance, reported by the percentage of correct and partially correct answers in rounds one and two, respectively, and further highlight the memory and compute requirements given by final tree size and LLM token cost at question time.
To systematically analyze the effect of each of our contributions, we compare \ours with different variants and ablations as detailed in \cref{sec:experiments,tab:orga:eval:baselines}.
\definecolor{relevantblue}{HTML}{053CDF}
\definecolor{forgottenred}{HTML}{941100}
\definecolor{correctgreen}{HTML}{008F00}
\definecolor{partiallycorrectorange}{HTML}{608000}
\definecolor{wrongred}{HTML}{FF3300}
\newcommand{\examplefontsize}{\fontsize{6.5}{7}\selectfont}
\newcommand{\correctsample}[1]{#1 \textcolor{correctgreen}{\quad\small\cmark}}
\newcommand{\partialsample}[2]{#1 \textcolor{partiallycorrectorange}{(\cmark)\ (\textit{#2})}}
\newcommand{\wrongsample}[2]{#1 \textcolor{wrongred}{\xmark\ \textit{(#2)}}}
\newcommand{\forgotsample}[1]{#1 \ \raisebox{-3pt}{\includegraphics[height=1em]{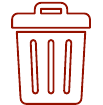}}\:\!}

\begin{figure}
    \centering
    \begin{subfigure}{1.0\linewidth}
        \centering
        \includegraphics[width=0.95\linewidth]{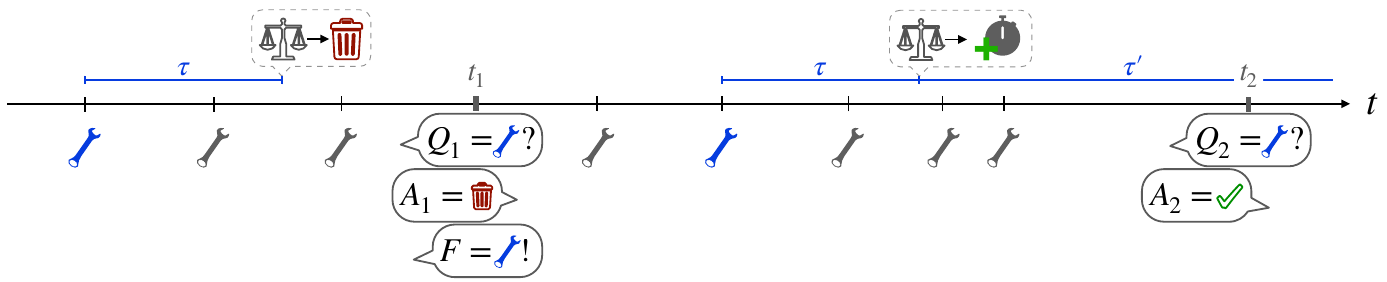}
        \\[-0.1cm]
        \caption{Evaluation procedure. The agent performs many actions (\raisebox{-3pt}{\includegraphics[height=1.2em]{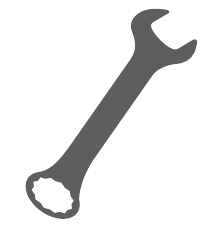}}\:\!) and is asked for details about some \textcolor{relevantblue}{relevant event} at a later point ($Q_1$). As the default lifetime $\tau$ of such event has already passed and therefore details are likely \textcolor{forgottenred}{forgotten} ($A_1$), the agent receives feedback ($F$), reinforcing the relevance. When asked later about a similar event ($Q_2$), the question should be answered \textcolor{correctgreen}{correctly} ($A_2$) since relevance estimation learned to extend the lifetime ($\tau'$).}
        \label{fig:eval:procedure}
    \end{subfigure}
    \\[0.2cm]
    \begin{subfigure}{0.49\linewidth} 
        \centering
        \examplefontsize
        \TikzTwoRoundEmvExample{
            03/20 10:53
        }{
            To where did you first transport a knife?
        }{
            At the countertop
        }{
             \forgotsample{There is no record of me transporting a knife in my available history.}
        }{
            You should always remember the location to where you transport a knife
        }{
            03/26 12:26
        }{
            To where did you last transport a knife?
        }{
            At the countertop
        }{
             \correctsample{The last place I transported a knife to was the countertop.}
        }
        \caption{Example from TEACh simulation (successful relevance learning and retrieval)}
        \label{fig:eval:ex1}
    \end{subfigure}
    \begin{subfigure}{0.49\linewidth}
        \centering
        \examplefontsize
        \TikzTwoRoundEmvExample{
            06/04 20:40
        }{
            When did you first meet Leonard?
        }{
            At 2025/06/04, 10:53:22
        }{
            \wrongsample{I first met Leonard on June 4, 2025, at around 11:54 AM.}{correct: 10:53}
        }{
            You should always remember the exact time when you meet someone
        }{
            06/05 12:21
        }{
            When did you last see Joana?
        }{
            At 2025/06/04, 21:38:39
        }{
            \partialsample{The last time I saw Joana was on June 4, 2025, at 21:30.}{correct: 21:38}
        }
        \caption{Example from \armarVII (successful relevance learning, retrieval slightly off)}
        \label{fig:eval:ex2}
    \end{subfigure}
    \begin{subfigure}{0.49\linewidth}
        \centering
        \examplefontsize
        \TikzTwoRoundEmvExample{
            06/04 19:43
        }{
            Where did you first see the keys?
        }{
            The kitchen sink
        }{
            \forgotsample{I do not have a record of where I first saw the keys.}
        }{
            You should always remember the exact location where you see a valuable item like the keys
        }{
            06/05 12:21
        }{
            Where did you last see the wallet?
        }{
            On the mobile kitchen counter
        }{
            \partialsample{The wallet was last seen on June 4, 2025, between 21:23 and 21:35, near the kitchen counter.}{correct: 21:40}
        }
        \caption{Example from \armarVII (correct relevance generalization, retrieval slightly off)}
        \label{fig:eval:ex3}
    \end{subfigure}
    \begin{subfigure}{0.49\linewidth}
        \centering
        \examplefontsize
        \TikzTwoRoundEmvExample{
            04/29 14:25
        }{
            What object did you load into the dishwasher the first time you did the task?
        }{
            The red plate
        }{
            \forgotsample{I have already forgotten...}
        }{
            You should always remember which object you \textbf{load} into the dishwasher when you do the task
        }{
            05/03 18:54
        }{
            What object did you \textbf{unload} from the dishwasher the last time you did the task?
        }{
            The red plate
        }{
             \forgotsample{I already forgot what object I unloaded from the dishwasher the last time I did the task.}
        }
        \caption{Example from \armarVII (failure of relevance learning)}
        \label{fig:eval:ex4}
    \end{subfigure}
    
    \caption{Overview of our evaluation procedure with model output examples}
    \label{fig:eval}
\end{figure}

\resultsSec{Forgetting \& \term{online memory construction} impose performance–efficiency tradeoffs}
Both forgetting and online memory construction negatively affect QA performance while reducing compute and/or storage requirements, each introducing a tradeoff.

Starting with forgetting, the negative impact on QA performance is especially visible in first-round success rate:
Comparing \term{online history tree construction} approaches in \cref{fig:results}, the first-round performance of systems with forgetting (\varA, \varB, \varD) is roughly half that of the non-forgetting system (\varF).
This is expected, as forgetting removes information that may be relevant for future QA.
These differences are even more pronounced for \term{offline history tree construction} (\varG versus \varJ) and the flat baselines (\varK versus \varL).
In contrast, forgetting naturally reduces memory and compute requirements --- final tree size of systems with forgetting is smaller, and QA token costs also grow more slowly with increasing history length (comparing $|h|=5$ and $|h|=25$).

Similarly, \term{online history tree construction} without forgetting performs worse than the offline system (\varF versus \varJ).
Regarding memory requirements, each offline system produces a smaller final tree than its online counterpart (\varA versus \varG, \varF versus \varJ).
This is expected, as the offline system has the advantage of access to the full retrospective event history.
On the contrary, \term{online history tree construction} exhibits lower QA token costs, as it shifts \term{grouping and summarization} effort away from question time.

\resultsSec{\term{Feedback-based relevance learning} improves round two}
\term{Feedback-based relevance learning} causes notable improvements from the first to the second round, whereas methods without \term{relevance learning} stagnate or decline across rounds.
For instance, for both longer and shorter histories, \varA and \varG improve from round one to round two, while \varB and \varJ decrease.
This can be explained by the fact that approaches without \term{relevance learning} suffer from the growing amount of information over time, as the second-round QA process has access to more memories, which increases the likelihood of false retrieval.
In contrast, methods with \term{relevance learning} can prioritize retained information as well as tree summaries, thereby effectively benefiting from feedback.

For $|h|=5$, \term{relevance estimation with feedback-based relevance learning} can mostly compensate for the negative effects of forgetting in the second question round.
Specifically, the second-round performance of \varA is comparable to that of \varF, despite substantially smaller average and final tree sizes.

\resultsSec{Interaction of relevance estimation \& \term{feedback-based relevance learning}}
The effectiveness of relevance estimation is two-fold:
On the one hand, we can see that \term{relevance estimation without feedback-based relevance learning} is rather ineffective.
Without a learned definition of relevance, deciding which events are relevant and which can be forgotten becomes arbitrary.
On the other hand, relevance estimation combined with \term{feedback-based relevance learning} has a synergetic effect, improving performance from round one to round two while effectively managing storage and compute requirements.

Specifically, comparing the variants without \term{relevance learning} with and without relevance estimation (\varB and \varD), we observe very similar behavior, both declining from round one to round two.
In contrast, our full system (\varA) achieves the largest gain in performance from first to second round while also maintaining low tree size and QA token costs.
In \cref{app:results}, we additionally ablate relevance estimation while keeping \term{feedback-based relevance learning} that is then only applied to summarization (\varC).
In the simulated evaluation, this variant performs comparably to our full system and also improves from round one to round two.
However, its performance declines in the real-world experiments, showing that relevance estimation is an important component for more diverse and realistic input data.
We further remove \term{relevance learning} from summarization while keeping it only for relevance estimation during forgetting (\varAA, \cref{app:results}).
This variant performs worse than the full system (\varA) in round two.
Overall, these results indicate that applying \term{feedback-based relevance learning} to both forgetting and summarization is beneficial.

\resultsSec{Interaction of \term{online memory construction} \& forgetting}
Our results suggest that \ours's forgetting mechanism is more compatible with \term{online history tree construction} than with offline-built or flat histories.
Considering QA performance of systems with forgetting, variants with offline or flat history construction perform notably worse than the online counterpart.
While \term{offline history tree construction} has the advantage of producing more compact trees, this does not translate into lower question-time compute requirements due to the additional tree construction effort.
In contrast, QA performance of systems without forgetting is higher for offline and flat histories than for \term{online history tree construction}.
However, these approaches also come with higher question-time compute costs.

In particular, comparing \term{online and offline history tree construction}, forgetting severely impacts performance in the offline setting (\varG versus \varJ), while performance remains higher in the online variants despite forgetting (\varA to \varD versus \varF).
To explain this, we hypothesize that both forgetting and \term{online history tree construction} introduce a recency bias and are thus more compatible.
Moreover, \term{online history tree construction} tends to build deep hierarchies, so that higher-level summaries of past events remain available, whereas offline-built trees are severely affected by first-round time-based forgetting (\varG first round performance is near zero).
This effect can be partially mitigated in the second round through \term{relevance estimation with feedback-based relevance learning}; however, performance still does not reach that of \term{online history tree construction} systems with forgetting (\varG versus \varA).

\resultsSec{Scaling to long histories}
\ours successfully scales to long histories, improving through \term{feedback-based relevance learning} while keeping storage and compute requirements low.
However, the overall performance is lower (\cref{fig:results:qa_teach_25} versus \cref{fig:results:qa_teach_5}), which can be attributed to the increased difficulty of the QA task when the sequence of episodic events becomes longer.
On the contrary, the differences in tree size and QA token costs between methods become more pronounced.
Among the \term{online history tree construction} systems with forgetting, our full system (\varA) shows the largest improvement from the first to the second round.

\resultsSec{Non-hierarchical baselines}
Our results show that \term{relevance-based forgetting} with \term{feedback-based relevance learning} is very ineffective without the hierarchical structure of EM (\varK).
Since there is no summarization hierarchy and events are not evaluated in context of the current task when estimating relevance during forgetting, the learned relevance rules do not apply well.
In contrast, the non-forgetting, flat baseline acts similarly to an oracle, achieving very high QA performance (\varL).
However, due to the absence of a hierarchy, both flat variants suffer from exploding compute costs at question time when scaling to longer histories.

\resultsSec{Transfer to real-world robot recordings}
\ours (\varA) successfully transfers to real-world recordings from \armarVII.
Due to the increased complexity and diversity of real-world EM content and more realistic human-curated QA samples, some effects become more pronounced:
As seen in \cref{fig:results:armarx}, \ours is the only system showing a clear improvement in QA performance from round one to round two (\varA), while the other forgetting systems stagnate (\varB, \varD) and the non-forgetting systems substantially decline (\varF, \varJ), despite higher memory and compute costs.
While the \varJ system appears to perform best at first glance, its performance decreases notably from round one to round two.
In addition, it incurs higher token costs and already exceeds \varA in $N_f$, despite having access to full \term{offline knowledge during history tree construction}.

Qualitative success and failure examples can be found in \cref{fig:eval:ex1,fig:eval:ex2,fig:eval:ex3,fig:eval:ex4,app:samples:realworld}.
Possible failure cases include erroneous \term{relevance learning}, relevance estimation missing relevant items, and QA retrieval errors, where the relevant information is available at query time but not correctly retrieved from the history tree.
Qualitatively, most failures of \ours fall into the latter category, confirming the findings of \cite{barmann_episodic_2025} that retrieval is a major source of errors.
With respect to relevance learning, we observe multiple instances of successfully generalized behavior, for instance regarding \enquote{valuable items} as shown in \cref{fig:eval:ex3}.

\resultsSec{Real-world humanoid robot deployment}
We deploy \ours to run live on the humanoid robot \armarVII.
For this, \term{online history tree construction} runs asynchronously, while forgetting is executed nightly and on‑the‑fly when the update queue is empty.
EMV always uses a copy of the most recent finished history tree update, thus being able to answer the question without waiting for tree construction.
More implementation details in \cref{sec:methods:real-world-demo}.

\begin{figure}
    \centering
    \begin{subfigure}{0.95\linewidth}
        \centering
        \includegraphics[width=\linewidth]{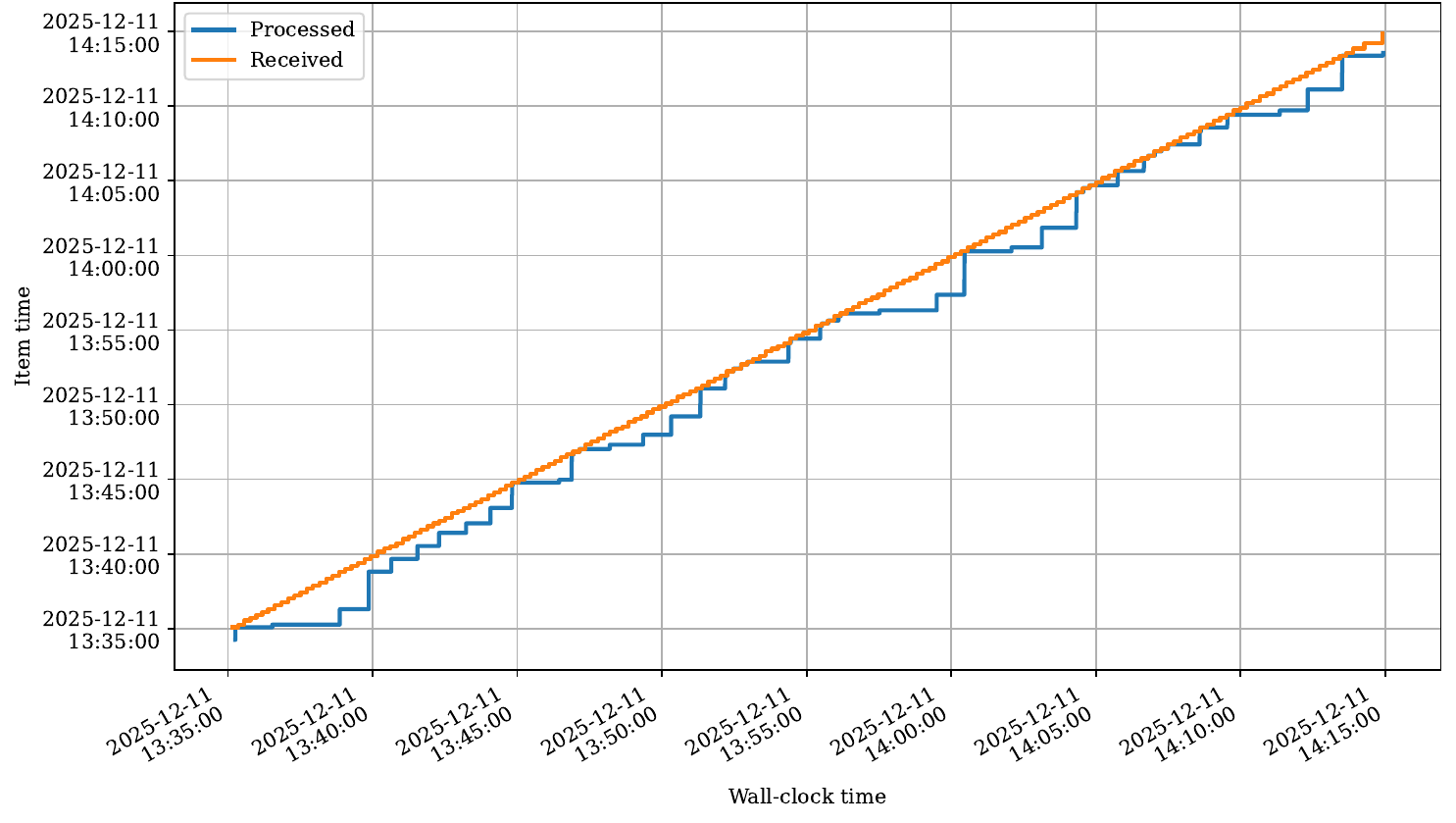}
        \caption{Visualization of the \term{online history tree construction} progress over time. The difference between Received and Processed is the delay introduced by the \term{incremental grouping and summarization algorithm}, which depends on the type of actions performed and observations made by the robot.}
        \label{fig:results:deploy:progress}
    \end{subfigure}
    \begin{subfigure}{0.95\linewidth}
        \centering
        \includegraphics[width=\linewidth]{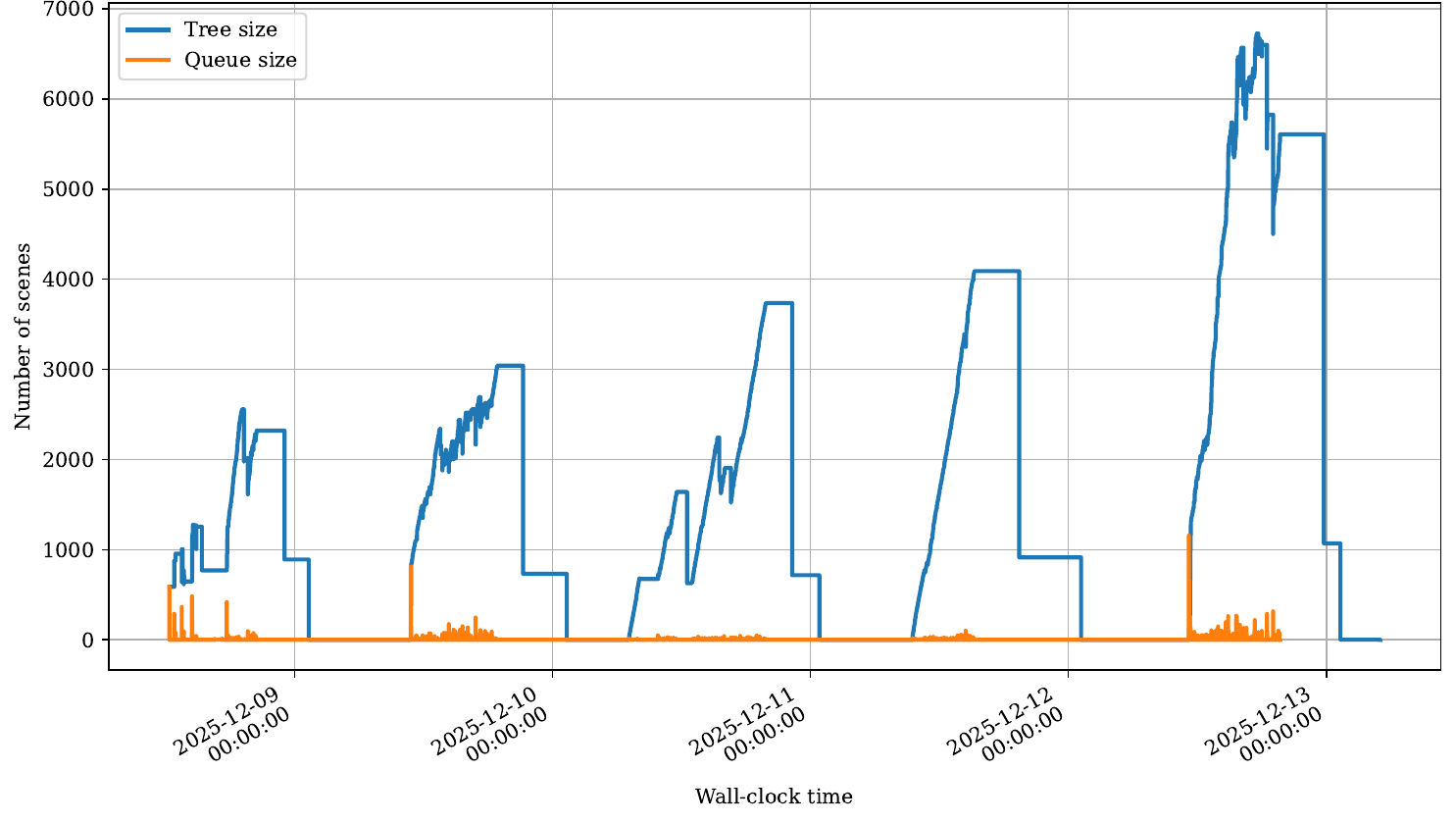}
        \caption{Number of queued items and tree size (measured by lowest-level scene graph nodes) over multiple days of live deployment of \ours. While the tree grows throughout each day (with forgetting only slightly reducing its size during periods of inactivity), forgetting is scheduled for nighttime and removes most irrelevant leaf nodes.}
        \label{fig:results:deploy:treesize}
    \end{subfigure}
    \caption{\ours deployed live on \armarVII}
    \label{fig:results:deploy}
\end{figure}

We ran this deployment of \ours for multiple weeks to qualitatively assess system performance.
On average, the delay introduced by \term{online history tree construction} during active periods is $47 \pm 41$ seconds, with large variances caused by differences in robot activity:
When there is a high density of incoming information, \term{online history tree construction} updates multiple levels of the history tree, resulting in higher delays (maximum observed delay: $7.2$ minutes).
In contrast, monotonous or low-frequency activities only trigger fast low-level tree updates (tenth percentile: $3$ seconds).
\cref{fig:results:deploy:progress} illustrates the online system's behavior over a representative time window.
Forgetting does not introduce any additional delay due to its asynchronous and interruptible implementation, while still effectively controlling tree size and thus easing subsequent updates.
\cref{fig:results:deploy:treesize} shows the number of lowest-level tree nodes over multiple days.
Qualitative QA samples from the live deployment are provided in \cref{app:samples:live}.

\newpage
\thispagestyle{empty}
\clearpage


\section{Discussion}
We introduce \ours, a system for modeling and verbalizing EM of an egocentric agent, and demonstrate its applicability to a real-world humanoid robot.
Through \term{time-aware incremental grouping and summarization}, the system constructs a tree-based representation of EM that is updated online as new observations and actions occur.
Further, to efficiently limit EM storage and query-time compute requirements, \ours applies a \term{decay-based forgetting mechanism with LLM-based relevance estimation}, conditioned on a \term{relevance measure that is incrementally learned from task and user feedback}.

Our experiments validate the effectiveness of our contributions:
\term{Online history tree construction}, beyond its conceptual advantage over its offline variant, produces reasonable trees that facilitate QA about the robot's past, although at the cost of increased tree size compared to offline-built trees.
Our forgetting mechanism effectively prunes these trees to a reasonable size, resulting in a lower QA token budget than \term{offline history tree construction} (considering that tree construction is not performed at question time).
Finally, relevance estimation during forgetting, combined with \term{feedback-based relevance learning} for both forgetting and summarization, enables our system to improve over time.
In contrast, systems without \term{relevance learning} stagnate or degrade as they process more experiences.
Our approach transfers to real-world data, remains scalable to very long histories through its \term{relevance-based forgetting mechanism}, and consistently adapts to user tasks via \term{feedback-based relevance learning}.

Comparing our system with the non-forgetting baselines, these can be interpreted as an upper bound for the performance of the respective approach with forgetting --- however, this is a soft upper bound, since \term{relevance-based forgetting} can ideally even improve performance by simplifying EM retrieval.
We do observe such behavior in the experiments on recordings from \armarVII, where \ours outperforms the non-forgetting baseline in the second round.
Similarly, \term{offline history tree construction} without forgetting has an obvious advantage, especially in the first round, as it has access to the full retrospective tree at question time and thus solves a simpler task.
This is even more pronounced for the non-forgetting flat baseline, which relies directly on the list of recorded events for QA, avoiding additional processing steps that may introduce errors (\eg summarization or forgetting).
However, as shown in our experiments, systems without forgetting --- especially the flat variants without hierarchical summarization --- come with high memory and compute costs for longer histories.
This is consistent with the findings of \cite{barmann_episodic_2025}, where non-hierarchical baselines scaled to even longer histories lead to exploding token costs and eventually a performance drop once the LLM's effective context length is reached.
Therefore, such systems are not suitable for real-world scenarios involving life-long experience streams.

Comparing the costs of \ours with the baselines, the total compute costs of the full system are higher.
\term{Incremental grouping and summarization} frequently invokes the LLM on every observation, while \term{offline history tree construction} can directly access all information seen so far.
Similarly, relevance estimation during forgetting adds significant LLM inference costs, as multiple nodes may need to be inspected.
However, both processes can be performed in parallel with other tasks: \term{online history tree construction} is performed while the robot executes its actions, and forgetting can be deferred to periods of low workload (\eg at night).
Therefore, \ours has the lowest effort at question time, as the history \term{constructed online} can be accessed instantly, and forgetting reduces the amount of data that must be searched during QA.

Our deployment of \ours on \armarVII demonstrates that the system is applicable in real-world scenarios.
While we do not claim real-time capability due to delays introduced by LLM inference, the qualitative results show that these delays are manageable.
Moreover, most EMV questions do not concern the robot's most recent activities but rather past events, making minute-scale delays less critical in practice.
With future improvements in small but capable real-time LMs, \ours provides a framework for efficiently organizing and accessing EM.

Our system is naturally limited by the performance and inference cost of the underlying LLMs.
Since \term{online history tree construction} relies on \term{incremental grouping and summarization}, small differences in the input can propagate through the hierarchy and lead to substantially different outcomes.
Similarly, the quality of the foundation models inferring information from the raw input is crucial, as their outputs form the basis of the lower tree levels.
For instance, a misrecognized object at a lower level may propagate to higher-level summaries, especially if it is similar to an item previously learned to be relevant.
With respect to \term{relevance learning} itself, representing relevance as a set of natural language rules is inherently imprecise, and LLM-based modification of these rules can be prone to hallucinations or unintended deletions.
On the other hand, this representation enables easy inspection and manual correction when necessary.

In future work, our system can be extended along multiple axes:
To improve \term{feedback-based relevance learning}, multimodal signals should be incorporated, \eg visual user satisfaction \cite{kumagai_towards_2018} or pointing gestures \cite{constantin_multimodal_2023}.
Similarly, learned relevance rules can be applied to other modalities and system components, \eg by biasing object recognition toward important items to reduce the likelihood of missing them.
Further, extending \term{relevance learning} to multi-user scenarios is not trivial, as users may have differing or even conflicting definitions of relevance.
Concerning forgetting, relevance estimation could be further enriched. 
Inspired by cognitive science, relevance measures based on comparisons to previously observed memories could be used to prioritize surprising or unseen (novel) events for longer retention \cite{foster_role_2019,barto_novelty_2013}.
Moreover, human memory relevance is strongly influenced by their emotion \cite{tyng_influences_2017}, which could be transferred to robotics by combining task success, social cues, and other contextual factors.

\clearpage


\section{Methods}
\label{sec:methods}

\subsection[H2-EMV]{\oursInHeader}

\paragraph{Task and Architecture Overview}
We focus on the task of Episodic Memory Verbalization (EMV), \ie given a stream of observations, we construct a memory representation of the agent's past, and access it to answer natural language questions about the observed history at subsequent points in time.
To approach this problem, our system builds upon \textsc{H-Emv} \cite{barmann_episodic_2025}.
From the sequence of observations, \textsc{H-Emv} derives a \emph{history tree}, a hierarchical data structure to represent EM, by recursively grouping and summarizing events using an LLM.
Given a subsequent question, an agentic LLM is used to efficiently explore the history tree, iteratively expanding nodes until the question can be answered.

In this work, we extend \textsc{H-Emv} with three major contributions:
First, we transfer EM construction from an offline to an online setting, processing the stream of observations as they arrive (= online), rather than constructing a tree retrospectively from a long recording (= offline).
Second, we introduce \term{decay-based forgetting with relevance estimation}, in which history tree nodes that have reached their lifetime are inspected by an LLM to estimate a relevance factor that may prevent them from deletion.
Third, as relevance is highly task- and user-dependent, we introduce a mechanism for \term{incremental learning of a relevance measure from previous queries and user feedback}, influencing both summarization during history tree construction and relevance estimation during forgetting.
Each contribution is detailed in the following.

\paragraph{\term{Online History Tree Construction}}
We assume a timestamped stream of raw observations $(r_1, r_2, ...)$ with $r_i$ observed at time $t_i$.
In our context, a raw observation is a highly multimodal data snapshot, containing information from the robot's perception and proprioception. 
From this stream, our goal is to continuously build and update a \textit{history tree} $\mathcal{H}$ representing EM, \ie a hierarchical representation of the experiences or events that have occurred so far.
More specifically, following \cite{barmann_episodic_2025}, $\mathcal{H}$ is a tree with scene graph instants $s_i$ at the lowest level $\ell=1$, event summary nodes $e_j$ at level $\ell=2$, goal summary nodes $g_k$ at level $\ell=3$, and higher-level summary nodes $h_n$ for $\ell>3$.
A scene graph instant is directly associated with one raw observation and enriches it with additional inferred information, such as a set of detected objects.
Each (event, goal, or higher-level) summary node $n$ contains a natural language summary $S_n$ and a set of child nodes, $\text{children}(n)$, from the layer below.

Given a new observation $r_t$ at time $t$, we first deduce information from it to fill the scene graph instant $s_t$.
This procedure, as well as construction of the lower levels of the hierarchy ($\ell \leq 3$), differs across domains
(simulated agent and real robot)
and is further elaborated for each specific experiment in \cref{sec:experiments}.

On top of the predefined levels of the history tree, \cite{barmann_episodic_2025} apply a recursive grouping and summarization algorithm to construct higher-level summary nodes.
However, this algorithm operates by inspecting the complete set of nodes at a given level, \ie it is an offline procedure.
In this paper, we extend it to a more realistic \term{online setting by updating the history tree incrementally}.

Given a new node $n_t$ at time $t$ (at the highest predefined level of the tree), we aim to incorporate it into $\mathcal{H}_{t-1}$ to obtain $\mathcal{H}_t$.
This is a non-trivial process, as it is not immediately obvious at which level the new observation belongs to the ongoing actions, or at which level it initiates a new one (see \cref{fig:system:inc_tree}).
Moreover, the new observation may provide evidence about the ongoing action that requires revising previously made decisions.
For instance, when a user requests an action from a robot, it may initially appear to start a new task, but later turn out to be a continuation of a higher-level ongoing task given subsequent requests.

For this reason, we propose using an LLM for \term{incremental grouping and summarization}.
Broadly speaking, at a given layer $\ell$ in the tree, the LLM is prompted with the current grouping of items into summary nodes at level $\ell + 1$, as well as the new node $n_t$.
The LLM is additionally conditioned on a set of natural-language relevance rules learned from user feedback, as explained in \cref{sec:methods:dialog_learning}.
It then produces an updated grouping of items from level $\ell$ to new or revised summary nodes.
This process is iterated in a bottom-up manner until there is no change on any layer.
A full prompting example can be found in \cref{app:prompts:grouping}.

\begin{algorithm}[b]
\algsize
\caption{\term{Time-aware incremental grouping and summarization}}
\label{alg:orga:methods:time_merge}
\begin{algorithmic}[1]
    \Require existing parents $P_{old}$ on layer $\ell + 1$
    \Require new items $N_{new}$ on layer $\ell$ \Comment{no parent yet}

    \State $N_{all} \gets [\ n \in \operatorname{children}(p)\ |\ p \in P_{old} \ ]$
    \State $C \gets \operatorname{time-based-cluster}(N_{all})$

    \If{$|C| = |N_{all}|$}
        \State Set C to one single cluster \Comment{avoid single-item clusters}
    \EndIf

    \State $P_{merged} = [\ ]$
    \ForAll{cluster $c \in C$}
        \State $P_c \gets [\ p\in P_{old} \ |\ \operatorname{children}(p) \subseteq c \ ]$
        \State $N_{new,\,c} \gets N_{new}\, \cap\, c$
        \If{$N_{new} = \varnothing$}
            \State $P_{merged} \gets P_{merged}\ \circ\ P_c$
            \Comment{avoid re-summarizing}
        \Else
            \State $P_{merged} \gets P_{merged}\ \circ\ \operatorname{group-and-summarize}_{LLM}(P_c, N_{new,\,c}, \ell)$
        \EndIf
    \EndFor

    \Ensure new list of parents $P_{merged}$
\end{algorithmic}
\end{algorithm}

\begin{algorithm}[tb]
\algsize
\caption{\term{Online history tree construction}}
\label{alg:orga:methods:inc_tree}
\begin{algorithmic}[1]
    \Require new scene $s_t$ at time $t$
    \Require history tree state $\mathcal{H}_{t-1}$
    
    \State $\mathcal{T} \gets \operatorname{copy(\mathcal{H}_{t-1})}$ \Comment{working copy}
    \State $\ell = 0$
    \State $N_{new} \gets \{ s_t \}$
    \While{$\ell < L - 1$}
        \State $P_{old} \gets \operatorname{edge-children}(\mathcal{T}, \ell + 1)$

        \If {$\operatorname{prevent-push-to-upper-layer}(N_{new}, \ell)$}
            \State insert $N_{new}$ into latest parent
            \State update latest parent's summary using $LLM$
            \Break
        \EndIf

        \State $t_{cutoff} \gets \operatorname{visibility-cutoff}(\mathcal T, \ell, N_{new})$
        \State $P_{visible} \gets [\ p \in P_{old} \ | \ \operatorname{end-date}(p) \geq \ t_{cutoff} ]$
        \State $P_{merged} \gets \operatorname{time-aware-group-and-summarize}_{LLM}(P_{visible}, N_{new}, \ell)$ \Comment{\Cref{alg:orga:methods:time_merge}}
        \State $P_{new} \gets [\ p \in P_{old} \ | \ \operatorname{end-date}(p) < \ t_{cutoff} ] \ \circ \ P_{merged}$

        \State $i_{changed} \gets \operatorname{index-of-first-change}(P_{new}, P_{old})$
        \If {$i_{changed}$ is None}
            \Break
        \EndIf

        \State $N_{new} \gets P_{new}\,[\,i_{changed}\,:\,]$

        \State $\ell \gets \ell + 1$
        
        \If {$\ell = L - 1$} \Comment{last iteration}
            \State $\mathcal T \gets \operatorname{simple-summarize}_{LLM}(P_{new})$
        \Else
            \ForAll {$p_o \in P_{old}\,[\,i_{changed}\,:\,] $}
                \State siblings $S \gets \operatorname{children}(\operatorname{parent}(p_o))$
                \State remove $p_o$ from $\mathcal T$
                \State remove all $s \in S$ from $\mathcal T$
                \State $N_{new} \gets N_{new}\, \cup\, S$
            \EndFor
        \EndIf
        
    \EndWhile

    \State $\mathcal{H}_t \gets \mathcal T$

    \Ensure new history tree state $\mathcal{H}_t$
\end{algorithmic}
\end{algorithm}

While LLMs perform well at semantically grouping items that belong together, they often fail to properly account for the timing of actions.
This is especially relevant when deploying online systems in a life-long manner:
For instance, a robot may be turned off during the night.
Consequently, the stream of scenes may contain many smaller or larger gaps, where the signal before and after a gap may be identical (\eg the robot will be in the same place after the night), but the observations should not be grouped together in the tree to form a meaningful hierarchy (\eg one high-level summary node representing each day).
Therefore, we propose \textit{time-aware incremental grouping and summarization} (\cref{alg:orga:methods:time_merge}) that extends the LLM-based procedure described above with an algorithm based on the temporal statistics of the current tree state.
Specifically, instead of presenting all possible items to the LLM, we introduce time-based visibility cutoffs and clustering to prevent merging items that are temporally distant.
The full \term{online history tree construction} algorithm, which uses time-aware incremental grouping and summarization internally, is shown in \cref{alg:orga:methods:inc_tree}.
It additionally considers the distance of a new node to the most recent parts of the tree in relation to existing gaps between nodes at that level to prevent promoting certain items to a higher level if doing so would create unreasonable temporal spans.

\paragraph{\term{Relevance-Based Forgetting}}
To prevent the history tree from growing endlessly, we implement a forgetting mechanism that prioritizes \enquote{relevant} information and deletes irrelevant old memories.
Initially, upon creation of a node $n$, an expiration time is set as
$$\tau_n = t_\text{end} + \Delta t_\ell \cdot \gamma_\ell$$
with
$$\gamma_\ell = \begin{cases}
    1,          & \ell \leq 3 \\
    2^{\ell-3}, & \ell > 3 \quad \text{(higher-level nodes)}
\end{cases}$$
$t_\text{end}$ denotes the end of the timespan describing the events in a node, $\Delta t_\ell$ describes the default timespan depending on the level of the node (\eg 15 minutes for raw observations or one day for goal-level nodes) and the level $\ell$ is enumerated from 0 (for raw observations) to the maximum level minus one.

After each invocation of the \term{online history tree construction} algorithm, we recursively traverse all nodes in a top-down manner to check whether they have expired (see \cref{fig:system:forgetting}).
Specifically, at time $t_\text{now}$, each node $n$ is checked against the forgetting rule
$$\tau_n < t_\text{now}$$
If this is the case, a relevance factor $\alpha$ is estimated
$$\alpha = \operatorname{estimate-relevance}(n)$$
and the node's expiration time is extended by
$$\tau_n \gets \tau_n + \alpha \cdot \Delta t_\ell$$
If the extended expiration time still satisfies $\tau_n < t_\text{now}$, the node is forgotten.
For every non-expired node, the forgetting check is recursively applied to all children in $C = \operatorname{children}(n)$, and finally, the expiration time of $n$ is set to
$$\tau_n \gets \max\left(\max_{c \in C}\ \tau_c,\ \ \tau_n\right)$$
Parent items are assigned the maximum of their own and their children's expiration dates, ensuring that a parent never expires before all its children have expired.

To determine the relevance factor $\alpha$ of a node, we prompt an LLM with the current node, its parent node (as context), and the current set of natural language relevance rules, and ask it to produce a numeric relevance score $\alpha$, which acts as a multiplier to extend the node's lifetime.
The LLM may also return "inf" to indicate that an item should be kept indefinitely.
A prompting example can be found in \cref{app:prompts:estimation}.
Our proposed mechanism can be classified as an offline time-based decay mechanism with an importance factor using the definitions from \cite{plewnia_forgetting_2024}.

Forgotten tree nodes are replaced by \enquote{forgotten} placeholders that retain only the time range and the first line of the node's own summary, dropping all other content and children.
The summary is hidden for EMV but visible to the \term{grouping and summarization LLM} during \term{online history tree construction}, ensuring that the parent's summary is derived from the forgotten node's content rather than explicitly stating that information has been forgotten.
Adjacent placeholders are merged if necessary by concatenating their short summaries.
This way, the history tree explicitly indicates that information for certain time spans has been forgotten, while other information is retained.

\paragraph{Dialog Manager \& \term{Relevance Learning}}
\label{sec:methods:dialog_learning}
To answer questions about the robot's experiences, \cite{barmann_episodic_2025} propose using an LLM as an agent that interactively explores the history tree to retrieve the requested information and respond accordingly.
We use this EMV system as a subcomponent of our dialog manager.
Specifically, given a user utterance, the dialog manager decides whether it is a question that should be forwarded to the EMV module, user feedback that should be passed to the \term{feedback-based relevance learning} component, or another type of utterance that can be answered directly.
We implement the dialog manager using an LLM prompted in an interactive Python console format, following \cite{barmann_incremental_2024}, providing EMV and \term{relevance learning} as functions to invoke.
\term{Relevance learning} itself also employs an LLM to update the current set of natural-language relevance rules based on the new user feedback.
This LLM is free to add, update, or remove relevance rules to support rule generalization and correction.
A prompting example can be found in \cref{app:prompts:learning}.
The revised relevance rules subsequently influence both the summarization of items during \term{online history tree construction} as well as the relevance estimation during forgetting, thereby better aligning the system's behavior with the user's expectations.

\subsection{Experimental Setup}
\label{sec:experiments}

\subsubsection{Simulated Robotic Agent Recordings}
To quantitatively evaluate our proposed full system --- specifically incremental history building,
\term{relevance-based forgetting}, and \term{feedback-based relevance learning} --- we conduct robotic experiments on simulated episodes.

\paragraph{Dataset}
We use simulated episodes from TEACh \cite{padmakumar_teach_2022}.
Specifically, these recordings capture the observations, state, and dialog of a human-operated agent in a simulation environment receiving task instructions from another human via a chat interface.
We repurpose these episodes as robotic experience streams and concatenate multiple episodes with randomized plausible dates and times to form longer histories (following \cite{barmann_episodic_2025}).
To measure the impact of history length $|h|$, we generate histories consisting of $|h|=5$ and $25$ TEACh episodes.
Each history is then annotated with grammar-generated, timestamped QA pairs, \ie each question is associated with a specific point in time during the replay of actions in the history.
For instance, some questions are asked after completing two tasks on one day, while others are posed several days later after additional actions have been performed.

For the purpose of evaluating our mechanism for interactive relevance learning, we generate QA pairs designed to specifically test the system's ability to improve from feedback.
In particular, we first randomly select an action or object to ask a question about, which must appear at least twice in the history with a large temporal gap between occurrences.
After the first occurrence of the action or object plus a fixed offset greater than the default lifespan of low-level nodes containing information about such action or object, we place an initial question.
Similarly, after the final occurrence of the action or object plus the corresponding offset, we place a second question.
By construction, the first question likely to be answered incorrectly, as the relevant details have already been forgotten.
The second question therefore tests the system's ability to improve the EM forgetting mechanism based on the user's feedback for the first answer.

\paragraph{Evaluation Procedure}
\Cref{fig:eval:procedure} illustrates the two-round evaluation procedure.
For each scene from the robotic experience stream, the \term{history tree is updated incrementally}.
In case a question is annotated for the current timestamp, the experience processing is paused and the dialog manager is invoked.
It receives the question and produces an answer using the EMV module to find the relevant information.
After answering a question, a real-world user could provide feedback to the system.
To automate and objectify our evaluation, this feedback is directly generated by reformulating the question.
For instance, if the question asks for details about event \enquote{X}, the feedback takes the form \enquote{You should always remember X}.
We always provide feedback to the system, independent of its answer to the original question.
This is equivalent to a learning mechanism that prioritizes information based on whether it is queried, \ie a system that directly infers relevance from the questions it receives.
The feedback is then passed to the dialog manager, after which the experiment continues with processing the experience stream until all question annotations have been handled.
Consequently, the feedback can influence further processing of the history and answers to later questions about it.
Using this experimental procedure and by construction of the dataset, the system's response for the first set of questions is most likely to indicate that the requested details have already been forgotten, whereas the second set of questions can be answered correctly in case the feedback mechanism successfully learned from the first feedback round.
Each history is evaluated independently, \ie relevance learning is not transferred across different samples.

\paragraph{Method}
Following \cite{barmann_episodic_2025}, we construct history trees from the simulated TEACh recordings.
These include the robot's knowledge of its current action, dialog input and output, as well as visual observations.
Since the focus of this evaluation is on the EM mechanism rather than perception capabilities, the observations use the ground-truth (GT) object data from the simulation state.
To speed up \term{history tree construction}, event and goal nodes are not generated by the grouping LLM, but instead derived from the robot's intrinsic knowledge of its current action and goal.
Specifically, a new event node (L2) is created whenever the scene graph or the current action changes, or when dialog input is received.
A goal node at L3 collects event nodes until an \enquote{interaction} action occurs (\ie any action other than navigation).
Effectively, this overwrites the implementation of $\operatorname{group-and-summarize}_{LLM}$ in \cref{alg:orga:methods:time_merge} with the described behavior for $\ell \in \{2, 3\}$.
Higher-level nodes are summarized as specified in \cref{alg:orga:methods:inc_tree}.

\paragraph{Metrics}
Our primary goal is to successfully answer questions about the robot's past.
Therefore, we evaluate the percentage of semantically (partially) correct answers $S_c$ ($S_p$) as defined in \cite{barmann_episodic_2025}.
Specifically, an LLM is provided with the question, GT answer, and the model's hypothesis, and is asked to classify the response into several semantic categories ranging from \emph{correct} to \emph{no answer}.
The LLM is few-shot prompted with samples from a training set, and its classification accuracy is validated to be sufficient for EMV evaluation.
Specifically, we extend the classifier prompting from \cite{barmann_episodic_2025} with additional hand-written few-shot samples and base instructions to improve performance on our multi-round dataset.
We evaluate this on a test set of 200 manually labeled samples from our TEACh evaluation, achieving $92\,\%$ classification accuracy with GPT-5 as a judge.

While QA accuracy is the primary metric, any forgetting mechanism will inevitably lead to some decline in this measure.
However, forgetting also provides benefits in terms of storage and processing efficiency, which we quantify using the following metrics:
\begin{itemize}
    \item Final tree size $N_f$: the number of L3+ nodes in the tree in its final state
    \item Average tree size $N_\varnothing$: the number of L3+ nodes in the tree averaged over all steps
    \item EMV LLM token budget $C_{qa}$: the number of tokens used by the LLM to explore the history tree during QA
    \item Forgetting token budget $C_f$: the number of tokens used by the forgetting mechanism itself
\end{itemize}
The last metric does not represent a benefit, but rather the additional costs introduced by the forgetting mechanism.
However, such an offline forgetting mechanism \cite{plewnia_forgetting_2024} can be postponed to periods of low workload (\eg at night), whereas $C_{qa}$ is more critical, as it directly affects responsiveness to user questions.
For the offline history tree construction baselines, also $C_{qa}$ includes the cost of building the tree itself, as this must be performed at question time.
We report LLM token budgets rather than QA or forgetting runtime, as execution time is highly dependent on the specific hardware setup.

To measure the effect of \term{feedback-based relevance learning}, we exploit the two-round structure of our dataset.
In particular, we match the generated pairs of questions corresponding to the first and last occurrences and report the percentage of (partially) correct answers separately, as $S_c^1$, $S_p^1$ and $S_c^2$, $S_p^2$, respectively.

\paragraph{Baselines}
As shown in \cref{tab:orga:eval:baselines}, we systematically ablate the components of our system to validate their desired effect.
Our full system (\varA) \term{constructs a history tree from the experience stream in an online manner}, uses the \term{decay-based forgetting mechanism with relevance estimation} to prolong the lifetime of important items, and takes in user feedback to learn improved definitions of what is relevant for forgetting and summarization.
The \varB variant removes \term{relevance learning}, thus relying solely on the LLM's common sense for summarization and relevance estimation during forgetting.
Further, the \varD variant removes relevance estimation altogether, using only time-based decay, \ie no item's lifetime is ever extended.
The incremental baseline (\varF) applies only \term{online history tree construction}, without forgetting or feedback-based relevance learning.

Additional baselines (\varG, \varJ) apply \term{offline} instead of \term{online history tree construction}.
In these variants, the complete recording of all experiences up to a given point is processed at question time to construct the history tree.
The \varG approach involves forgetting: after constructing the tree using the full history, the forgetting mechanism is applied prior to QA to remove expired events (that are not considered relevant by relevance estimation).
The \varJ variant is equivalent to the system of \cite{barmann_episodic_2025}, \ie \term{offline history tree construction} without forgetting or \term{feedback-based relevance learning}.

For further comparison, we introduce two additional baselines similar to the L3-variant of \cite{barmann_episodic_2025}.
Specifically, the \varK and \varL approaches do not build a history tree by recursive grouping and summarization, but instead solely operate on the predefined goal-level nodes.
These nodes are presented to the EMV LLM as a flat, expandable list, \ie the LLM still follows an agentic approach (including semantic search), but without a deep hierarchy, making it comparable to a multi-round RAG system.
The \varK variant applies \term{relevance-based forgetting with feedback-based relevance learning} in this setting, whereas \varL is equivalent to the L3 baseline of \cite{barmann_episodic_2025}.

\subsubsection{Real-World Humanoid Robot Recordings}
To validate \ours on real-world data, we apply it to recordings from our humanoid robot \armarVII \cite{asfour_armar_2017}.
While the sample size for quantitative evaluation is smaller than in the simulated evaluation setting, the goal is to assess whether the system transfers to real-world data including more realistic QA scenarios.

\paragraph{Data}
Using the ArmarX memory system \cite{Peller-Konrad2023MemorySystemRobot}, we recorded a set of episodes of \armarVII performing household tasks, giving robot demonstrations and interacting with users in our lab kitchen environment.
Including recordings from previous studies, the full dataset consists of 35 recordings totaling 20.5 hours of multimodal data, including images, robot proprioception, symbolic scene and action knowledge, and user interaction transcripts.
To construct even longer histories, analogous to the simulated evaluation in the previous section, we randomly concatenate multiple episodes with randomized dates and times, ensuring that similar events occur at least twice with a sufficient temporal distance (for \term{feedback-based relevance learning}).
To create the QA annotations, we first generate a large number of first-/second-round QA pairs as described above, using technical, unnatural vocabulary (\eg \enquote{When did you last NavigateToNamedLocation(location=counter)?}).
These samples are then manually filtered to retain only interesting and realistic scenarios, resulting in five histories with ten episodes each, annotated with ten QA pairs (\ie 20 QA samples per history).
Subsequently, the utterances are refined by an LLM to make them more natural (\eg \enquote{When did you last move to the counter?}), and finally manually reviewed and corrected to prevent errors introduced during LLM rephrasing.

\paragraph{Method}
From the multimodal stream of events, we construct lower-level history tree nodes following \cite{barmann_episodic_2025,liu_reflect_2023}.
L1 scene graph instants are created based on the symbolic scene information available in the working memory.
Event-level nodes (L2) are triggered based on changing scene graphs, user speech, or skill status updates (\ie new or finished actions).
In contrast to the TEACh simulation experiments, L3 nodes (representing goals) can be nested to represent subgoals pursued during a higher-level goal.
This is automatically derived from the data by using the skill invocation stack; \eg if the skill \enquote{BringObjectToHuman} invokes \enquote{GraspObject}, the latter is interpreted as a subskill of the former.
When applying the \term{online history tree construction} from \cref{alg:orga:methods:inc_tree}, this L3 nesting is fully transparent to the algorithm:
It treats all (potentially nested) L2 nodes as direct children of the high-level L3 goal, and the subgoal hierarchy is locally reconstructed upon changes.
This allows us to apply our algorithm without domain-specific modifications.

Following the notation from \cref{tab:orga:eval:baselines}, we only apply a subset of ablations to investigate whether the observations from simulation transfer to our real-world data.
Specifically, we test the full system (\varA), the variant without relevance learning (\varB), simple time-based decay forgetting (\varD), and the \varF baseline with \term{online history tree construction}, as well as the offline system using \term{relevance-based forgetting with feedback-based relevance learning} (\varG) and the \varJ baseline.

\paragraph{Metrics}
The reported metrics are equivalent to those used in the TEACh experiments above.
However, the classification of answers into semantic categories (affecting $S_c$, $S_p$) is only pre-annotated by an LLM and then manually corrected for every sample.
This procedure is necessary because the LLM classifier is few-shot prompted with TEACh samples and therefore exhibits lower accuracy under the real-world domain shift.
Moreover, we did not have access to sufficient independent real-world data to construct an adequately large labeled set for improving the classifier.

\subsection{Real-World Humanoid Robot Deployment}
\label{sec:methods:real-world-demo}
We deploy our system to run live on the humanoid robot \armarVII.
The method is equivalent to the experiments with \armarVII recordings described above, with one modification:
online history tree construction (\cref{alg:orga:methods:inc_tree}) can process a batch of scenes instead of a single scene, improving efficiency by handling multiple scenes collected since the last tree update.
For the technical deployment, scene construction and EMV are executed on the robot's internal PC, while the \term{online history tree construction} and forgetting processes are offloaded to a GPU server within the university network to enable parallel processing.
Specifically, upon each relevant trigger, scenes are constructed from the robot's memory, covering the interval from the last tree update to the current time.
Relevant triggers include the occurrence of a new skill event (\ie when a new action starts or ends), a speech event from either the robot or the user, or the recognition of a human face.
The resulting scenes are sent to the server, where \term{online history tree construction} runs asynchronously.
The server collects scenes from multiple updates in a queue while the \term{incremental grouping and summarization} algorithm is still handling a previous batch.
In contrast to the experiments above, forgetting is not necessarily triggered after each tree update, but is deferred when new updates are already queued.
If the queue is empty, forgetting is applied after a tree update.
The forgetting process is designed to be interruptible: tree recursion is stopped gracefully if a new concurrent update arrives, giving priority to updating the tree over forgetting.
In addition, forgetting is scheduled during periods of inactivity at night to maintain a compact tree for subsequent operation on the next day.

When receiving a question, the robot retrieves the latest available history tree from the server.
Ongoing updates do not block retrieval, as the system always accesses the most recent completed snapshot.
The EMV module is then applied to this tree.
The dialog manager is directly integrated into the LLM-based dialog system running on the robot \cite{barmann_incremental_2024}.
Specifically, in contrast to the quantitative experiments above, where the dialog manager selects among a limited set of three actions, the full dialog system operates as a general LLM agent with access to a rich API for the robot's perception and action.
We extend this API with two functions ---\texttt{answer\_question\_about\_my\_past}, \texttt{handle\_forgetting\_feedback} --- to enable the type of dialogs relevant to our study.

\subsection{Related Work}
\input{content/1a_related_work}

%% file: content/1a_related_work.tex
\paragraph{Episodic Memory in Robotics}
The concept of episodic memory (EM), sometimes also called autobiographical memory, was first introduced in human cognition by \cite{tulving_episodic_1972} and is nowadays applied in cognitive architectures and robotics, both to enable autonomous agent behavior and to better understand human EM \cite{prescott_synthesizing_2024}.
The most straightforward way to represent EM in artificial agents is to continuously keep track of what happens, and store relevant experiences in an explicit representation, such as a log-file \cite{zhu_autonomous_2017}, ontology \cite{beetz_knowrob_2018}, or a structured database \cite{petit_lifelong_2016,Peller-Konrad2023MemorySystemRobot}.
Other works explored latent vector representations of the past:
For instance, \cite{wang_modeling_2016,leconte_design_2016} use ART (adaptive resonance theory) networks to store and recall events and episodes. 
Training a neural network with an auto-encoder goal, \cite{rothfuss_deep_2018,Peller-Konrad2023MemorySystemRobot} create latent representations of past experiences, while \cite{barmann_deep_2021,dechant_search_2024} implicitly train the EM as part of end-to-end verbalization training (see below).
While such latent representations can provide more flexibility, they also lack interpretability, which is why a recent stream of works represents EM as text created by foundation models \cite{zeng_socratic_2023,liu_reflect_2023,long_seeing_2025}.
For instance, \cite{liu_reflect_2023} use three-layer text-based memory to reflect on failures during the robot's task executions, and \cite{long_seeing_2025} build a text-based knowledge graph to equip an agent with long-term memory.

\paragraph{Robot Experience Verbalization}
Verbalization of robot experience, a task initially proposed by \cite{rosenthal_verbalization_2016}, has been approached with various techniques:
Earlier systems paired explicit, log-style EM representations with grammar-based rules for verbalization of recorded experiences \cite{rosenthal_verbalization_2016,zhu_autonomous_2017}.
With the aim of overcoming templates and increasing flexibility, other works proposed to learn a model for robot experience verbalization end-to-end:
Both \cite{barmann_deep_2021} and \cite{dechant_learning_2023} use simulated robot data annotated with grammar-generated QA to train a transformer-based network to answer questions about a robot's past.
Since these systems struggle with generalization to unseen experiences and questions deviating from their training grammar, more recent works make use of foundation models to create training-free systems.
For instance, \cite{wang_i_2024,anwar_remembr_2025,katuwandeniya_what_2025,barmann_episodic_2025,alvarez-arias_connecting_2026} use existing vision models such as object detectors or vision-language models to annotate each scene.
From that, they create text-based representations of the robot's past, that are then queried by an LLM given a question or summarization request.
To facilitate efficient retrieval of past events, \cite{anwar_remembr_2025} use retrieval-augmented generation, while 
\cite{barmann_episodic_2025} prompt an LLM to dynamically explore a hierarchical representation of EM.
Balancing computational efficiency with response reliability, \cite{plewnia_combining_2025} introduce a hybrid system that combines rule-based memory querying with LLM-based question matching and template generation, reducing token consumption while maintaining high answer accuracy.
Focusing on verbalization of lower-level actions, \cite{wang_cori_2025} propose a system for communicating a robot's (intended) end effector trajectory without task-level knowledge by prompting a VLM with trajectory-augmented scene images.

Our work directly builds upon \cite{barmann_episodic_2025} using hierarchical representations created by foundation models, accessed by an LLM as agent for question answering.
However, it builds the hierarchical representation online instead of processing a recorded set of events retrospectively, and adds forgetting and feedback-based relevance learning.

\paragraph{Forgetting and Incremental Learning}
The implementation of forgetting in cognitive architectures has progressed from simple memory management (\cite{freedman_2011}, \cite{sigalas_2017}) to more biologically-inspired mechanisms. Plewnia et al. (2024)~\cite{plewnia_forgetting_2024} provide a comprehensive taxonomy of forgetting approaches for robotic episodic long-term memory.
One prominently used mechanism is time-based decay, which implements temporal degradation of memory traces and can be enhanced through importance factors that weight memories by relevance.
This allows systems to retain significant memories longer while discarding less important information, mirroring human memory consolidation.
Forgetting strategies differ along critical dimensions. 
Online methods operate during active system execution, enabling continuous memory management but potentially impacting performance, while offline methods perform maintenance during dedicated phases, allowing more intensive processing without interference. Additionally, periodic forgetting conducts scheduled memory reviews at fixed intervals \cite{sigalas_2017}, whereas on-access forgetting triggers updates during retrieval attempts, offering more context-sensitive retention \cite{freedman_2011}.
These mechanisms enable cognitive architectures to maintain computational efficiency while exhibiting adaptive, human-like memory behavior in long-term operational scenarios.
Our work differs from previous systems by proposing a new way of calculating the importance factor by an LLM, estimating the value of memories given an incrementally learned measure of relevance.

To incrementally learn from natural-language feedback, previous works have augmented LLMs with memories to update from feedback and retrieve when solving a new problem instance.
For instance, \cite{barmann_incremental_2024} use a memory of code-based interaction examples to incrementally learn new high-level behavior of a humanoid robot.
Similarly, HELPER \cite{sarch_open-ended_2023} use an LLM to create robotic task plans and update a language-program-memory with successful executions to reinforce correct behavior.
DROC \cite{zha_distilling_2024} extract task and skill knowledge from robot execution traces influenced by a natural-language correction handler, with high-level task constraints being represented in natural language and low-level skill parameters on a subsymbolic level.
In our work, we also keep a memory of natural-language relevance rules and dynamically update the memory using an LLM given new user feedback.
However, we do not apply it to control high-level robot behavior, but for steering the forgetting process of our EM.

%% file: content/2_appendix.tex
\section{Further Results}
\label{app:results}

In addition to the experiments explained in \cref{sec:experiments}, we performed additional ablation studies to identify the effect of each component systematically.
An overview of all performed experiments in show in \cref{tab:orga:eval:all_ablations}.

To analyze the effect of relevance learning on summarization in online tree building further, we constructed the variants \varAA, \varC and \varE.
Approach \varAA does not propagate the learned relevance rules to hierarchical summarization, applying it to relevance estimation during forgetting only.
In contrast, variant \varC does only use time-based decay for forgetting, \ie no item's lifetime is prolonged by relevance estimation, but the feedback learning is still used for the purpose of improving the hierarchical summarization.
Similarly, variant \varE drops the forgetting mechanism altogether while still keeping feedback learning for the summaries.
In offline tree construction, we further experimented with \varH removing the relevance learning, and approach \varI removing relevance estimation from forgetting. 

\begin{table}[ht]
    \centering
    \small
    \setlength{\tabcolsep}{1mm}
    \begin{tabular}{rr|c|cc|cc}
    &            & tree construction  &  \twocolb{forgetting} & \twocol{relevance learning} \\
    &            &                    &   time & relevance    & for forg. & for summ. \\\hline
    & \varA      &          inc       & \cmark &    \cmark    & \cmark    & \cmark       \\
  * & \varAA     &          inc       & \cmark &    \cmark    & \cmark    & \xmark       \\
    & \varB      &          inc       & \cmark &    \cmark    & \xmark    & \xmark      \\
  * & \varC      &          inc       & \cmark &    \xmark    & \xmark    & \cmark       \\
    & \varD      &          inc       & \cmark &    \xmark    & \xmark    & \xmark       \\
  * & \varE      &          inc       & \xmark &              &           & \cmark       \\
    & \varF      &          inc       & \xmark &              &           & \xmark       \\\hline 
    & \varG      &      offline       & \cmark &    \cmark    & \cmark    & \cmark       \\
  * & \varH      &      offline       & \cmark &    \cmark    & \xmark    & \xmark       \\
  * & \varI      &      offline       & \cmark &    \xmark    & \xmark    & \xmark       \\
    & \varJ      &      offline       & \xmark &              &           & \xmark       \\\hline
    & \varK      &   flat (L3)        & \cmark &    \cmark    & \cmark    & \cmark       \\
    & \varL      &   flat (L3)        & \xmark &              &           & \xmark       \\
    \end{tabular}
    \caption{Full list of ablation experiments. * marks the experiments that are not contained in \cref{sec:experiments}.}
    \label{tab:orga:eval:all_ablations}
\end{table}

The detailed results of all experiments can be seen in \cref{tab:orga:eval:results_5,tab:orga:eval:results_25,tab:orga:eval:results_armar}, visualized in \cref{fig:app:results}.
Next to the metrics defined in \cref{sec:experiments}, this also reports the following numbers:
\begin{itemize}
    \item[$S_c$] The percentage of correct answers, averaged over all QA samples (first and second round mixed)
    \item[$S_p$] The percentage of partially correct answers, averaged over all QA samples (first and second round mixed)
    \item[$S_\uparrow$] The percentage of QA sample pairs that improved from first to second round (wrong $\to$ partially correct or correct, partially correct $\to$ correct)
    \item[$S_\equiv$] The percentage of QA sample pairs that scored in the same category (correct, partially correct, wrong) in round one and two
\end{itemize}

Further, \cref{tab:orga:eval:forgotten_ratio} reports the \emph{forgotten ratio}:
Given a question, during the QA generation process, we know the target time span of events this question refers to. 
At question time, we then check the history tree whether all leaf nodes in the given reference time span are forgotten.
The forgotten ratio is the percentage of QA samples where this applies, \ie it counts the samples where the low-level information referred to in the question is already forgotten.
Note that the question could still be answered correctly if the necessary information is contained in higher-level summary nodes.
By design, the methods with forgetting have a forgotten ratio of $100 \%$ in round one, as the relevance of the asked events cannot be known in advance.
Only \varA, \varAA, \varG and \varK, which all include relevance learning from feedback and relevance estimation during forgetting, are able to obtain a forgotten ratio below $100 \%$ in round two (expect the non-forgetting baselines, which naturally forget $0 \%$).

\begin{table*}[ht]
    \centering
    \footnotesize
    \setlength{\tabcolsep}{3.5pt}
    \begin{tabular}{r|rrrrrrrr|rr|rr}
          & \multicolumn{8}{|c}{QA performance $[\%]$} & \multicolumn{2}{|c}{Tree size (L3+)} & \multicolumn{2}{|c}{Token cost $[K]$} \\
          & $S_c$ & $S_p$ & $S_\uparrow$ & $S_\equiv$ & $S_c^1$ & $S_p^1$ & $S_c^2$ & $S_p^2$ & $N_f$ & $N_\varnothing$ & $C_{qa}$ & $C_f$ \\
        \hline 
        \varA        
                & 27 & 41 & 34 & 50 & 20 & 34 & 34 & 48 &  52.2 & 38.8 &  9.4 & 152.9 \\
        \varAA  
                & 20 & 27 & 16 & 64 & 22 & 28 & 18 & 26 & 45.7 & 38.2 & 9.1 & 150.3 \\
        \varB   
                & 17 & 27 & 16 & 56 & 20 & 34 & 14 & 20 &  43.6 & 38.6 &  7.9 & 127.2 \\
        \varC      
                & 22 & 37 & 30 & 58 & 18 & 28 & 26 & 46 &  43.9 & 37.7 &  9.1 & -- \\
        \varD   
                & 17 & 26 & 18 & 58 & 18 & 30 & 16 & 22 &  42.7 & 37.7 &  8.3 & -- \\
        \varE   
                & 43 & 60 & 34 & 24 & 48 & 64 & 38 & 56 & 135.0 & 63.9 & 10.2 & -- \\
        \varF   
                & 38 & 56 & 20 & 38 & 48 & 66 & 28 & 46 & 131.8 & 63.1 &  9.1 & -- \\
        \hline  
        \varG   
                &  9 & 17 & 20 & 70 &  2 & 12 & 16 & 22 & 19.0 & 18.7 & 8.0 + 3.5 & 69.4 \\
        \varH   
                &  7 & 15 & 18 & 70 &  2 & 12 & 12 & 18 & 17.4 & 18.3 & 6.2 + 3.5 & 58.5 \\
        \varI   
                &  4 &  9 &  6 & 86 &  2 & 10 &  6 &  8 & 13.4 & 14.9 & 6.2 + 3.5 & -- \\
        \varJ   
                & 51 & 69 & 18 & 52 & 58 & 74 & 44 & 64 & 94.9 & 71.8 & 11.1 + 3.5 & -- \\
        \hline
        \varK 
                &  9 & 9  &  2 & 94 & 10 & 10 &  8 &  8 &   -- &   -- & 15.0 & 147.7 \\
        \varL 
                & 56 & 79 & 22 & 56 & 54 & 80 & 58 & 78 &   -- &   -- &  9.8 &   --  \\
    \end{tabular}
    \caption{
        TEACh experiments, $|h|=5$
    }
    \label{tab:orga:eval:results_5}
\end{table*}

\begin{table*}[ht]
    \centering
    \footnotesize
    \setlength{\tabcolsep}{3.5pt}
    \begin{tabular}{r|rrrrrrrr|rr|rr}
          & \multicolumn{8}{|c}{QA performance $[\%]$} & \multicolumn{2}{|c}{Tree size (L3+)} & \multicolumn{2}{|c}{Token cost $[K]$} \\
          & $S_c$ & $S_p$ & $S_\uparrow$ & $S_\equiv$ & $S_c^1$ & $S_p^1$ & $S_c^2$ & $S_p^2$ & $N_f$ & $N_\varnothing$ & $C_{qa}$ & $C_f$ \\
        \hline 
        \varA  & 17 & 22 & 20 & 64 & 14 & 20 & 20 & 24 & 265.1 & 148.3 & 10.2 & 1369.8 \\
        \varAA & 15 & 17 &  8 & 76 & 18 & 20 & 12 & 14 & 223.9 & 133.1 & 10.4 & 1354.7 \\
        \varB  & 23 & 27 & 14 & 66 & 24 & 32 & 22 & 22 & 178.5 & 114.4 & 8.8 & 1165.8 \\
        \varC  & 17 & 23 & 16 & 72 & 14 & 20 & 20 & 26 & 187.3 & 117.1 & 10.4 & -- \\
        \varD  & 19 & 22 & 10 & 70 & 22 & 28 & 16 & 16 & 175.3 & 112.6 & 9.0 & -- \\
        \varE  & 34 & 43 & 20 & 50 & 40 & 48 & 28 & 38 & 944.3 & 441.5 & 12.8 & -- \\
        \varF  & 24 & 31 & 22 & 48 & 26 & 34 & 22 & 28 & 917.8 & 431.3 & 11.1 & -- \\
        \hline
        \varG  & 10 & 16 & 20 & 74 &  4 & 10 & 16 & 22 & 26.5 & 26.9 & 9.9 + 5.3 & 363.5 \\
        \varH  &  7 & 13 &  8 & 78 &  8 & 16 &  6 & 10 & 29.1 & 28.9 & 8.3 + 5.3 & 324.7 \\
        \varI  &  3 &  8 &  8 & 88 &  2 &  6 &  4 & 10 & 22.7 & 21.8 & 8.1 + 5.3 & -- \\
        \varJ  & 32 & 38 & 10 & 58 & 42 & 48 & 22 & 28 & 621.6 & 353.6 & 15.7 + 5.3 & 0.0 \\
        \hline
        \varK  &  7 &  7 &  4 & 94 &  6 &  6 &  8 &  8 & -- & -- & 16.5 & 1330.4 \\
        \varL  & 71 & 80 & 18 & 64 & 70 & 82 & 72 & 78 & -- & -- & 37.0 & -- \\
    \end{tabular}
    \caption{
        TEACh experiments, $|h|=25$
    }
    \label{tab:orga:eval:results_25}
\end{table*}

\begin{table*}[ht]
    \centering
    \footnotesize
    \setlength{\tabcolsep}{3.5pt}
    \begin{tabular}{r|rr|rr}
          & \multicolumn{2}{|c}{val-unseen 5} & \multicolumn{2}{|c}{val-unseen 25} \\
          & round 1 & round 2 & round 1 & round 2 \\
        \hline 
        \varA & 100 & 64 & 100 & 80 \\
        \varAA & 100 & 70 & 100 & 80 \\
        \varB & 100 & 100 & 100 & 100\\
        \varC & 100 & 100 & 100 & 100 \\
        \varD & 100 & 100 & 100 & 100\\
        \varE & 0 & 0 & 0 & 0 \\
        \varF & 0 & 0 & 0 & 0 \\
        \hline
        \varG & 100 & 74 & 100 & 82  \\
        \varH & 100 & 100 & 100 & 100  \\
        \varI & 100 & 100 & 100 & 100 \\
        \varJ & 0 & 0 & 0 & 0  \\
        \hline
        \varK & 100 & 98 & 100 & 96 \\
        \varL &   0 &  0 &   0 &  0 \\
    \end{tabular}
    \caption{
        TEACh experiments, forgotten low-level ratio
    }
    \label{tab:orga:eval:forgotten_ratio}
\end{table*}

\begin{table*}[ht]
    \centering
    \footnotesize
    \setlength{\tabcolsep}{3.5pt}
    \begin{tabular}{r|rrrrrrrr|rr|rr}
          & \multicolumn{8}{|c}{QA performance $[\%]$} & \multicolumn{2}{|c}{Tree size (L3+)} & \multicolumn{2}{|c}{Token cost $[K]$} \\
          & $S_c$ & $S_p$ & $S_\uparrow$ & $S_\equiv$ & $S_c^1$ & $S_p^1$ & $S_c^2$ & $S_p^2$ & $N_f$ & $N_\varnothing$ & $C_{qa}$ & $C_f$ \\ 
        \hline 
        \varA & 16 & 36 & 34 & 54 & 12 & 24 & 20 & 48 & 231.8 & 185.5 & 13.9 & 3350.4 \\ 
        \varB & 11 & 24 & 20 & 58 &  8 & 26 & 14 & 22 & 205.6 & 173.6 & 15.0 & 2801.4 \\ 
        \varC & 16 & 23 & 14 & 64 & 18 & 26 & 14 & 20 & 207.0 & 176.8 & 18.0 & -- \\ 
        \varD & 11 & 22 & 14 & 68 & 10 & 24 & 12 & 20 & 193.4 & 171.6 & 16.3 & -- \\ 
        \varF & 29 & 46 &  6 & 62 & 36 & 56 & 22 & 36 & 459.6 & 226.0 &  9.7 & -- \\ 
        \hline
        \varG &  9 & 20 & 18 & 70 &  8 & 18 & 10 & 22 &  50.8 &  50.4 & 23.4 + 4.9 & 2255.3 \\ 
        \varJ & 38 & 53 & 28 & 50 & 32 & 54 & 44 & 52 & 259.2 & 162.1 & 13.7 + 4.9 & -- \\ 
    \end{tabular}
    \caption{
        \armarVII experiments
    }
    \label{tab:orga:eval:results_armar}
\end{table*}

\begin{figure*}[!ht]
    \centering
    \begin{subfigure}{1\textwidth}
        \centering
        \EvalDetailPlots[0.175][0.462]{85}{200}{40}{figures/data-teach-5.tsv}[6.4cm]
        \vspace{-0.4cm}
        \caption{TEACh $|h| = 5$}
        \label{fig:app:results:qa_teach_5}
    \end{subfigure}
    \\[0.3cm]
    \begin{subfigure}{1\textwidth}
        \centering
        \EvalDetailPlots[0.175][0.462]{85}{980}{40}{figures/data-teach-25.tsv}[6.4cm][nolegend]
        \vspace{-0.2cm}
        \caption{TEACh $|h| = 25$ }
        \label{fig:app:results:qa_teach_25}
    \end{subfigure}
    \\[0.3cm]
    \begin{subfigure}{1\textwidth}
        \centering
        \EvalDetailPlots[0.287]{80}{500}{30}{figures/data-armar7.tsv}[4cm][nolegend]
        \vspace{-0.4cm}
        \caption{real-world robot recordings from \armarVII}
        \label{fig:app:results:armarx}
    \end{subfigure}
    \caption{All results from simulated and real-world robot experiments}
    \label{fig:app:results}
\end{figure*}

\clearpage

\section{Qualitative Samples}
\label{app:samples}

\subsection{\armarVII evaluation}
\label{app:samples:realworld}
\TwoRoundEmvExample[Successful example]{
    2024/04/24 20:23:34
}{
    When was the first time you visited the laundry area?
}{
    At 2024/04/24, 12:49:31
}{
    I have no record of ever visiting the laundry area.
}{
    You should always remember the exact time when you visit the laundry area
}{
    2024/05/03 18:54:42
}{
    When was the last time you visited the laundry area?
}{
    At 2024/05/02, 16:19:57
}{
    May 2, 2024, at 16:19 was the last time I visited the laundry area.
}



\TwoRoundEmvExample[
    Partially correct example (GT missed by 8 minutes). 
    Note that the feedback is a more general rule than the first round question.
]{
    2025/06/04 20:40:38
}{
    When did you first meet Leonard Baermann?
}{
    At 2025/06/04, 10:53:22
}{
    I first met Leonard Baermann on June 4, 2025, at around 11:54 AM.
}{
    You should always remember the exact time when you meet someone
}{
    2025/06/05 12:21:13
}{
    When did you last see Joana Plewnia?
}{
    At 2025/06/04, 21:38:39
}{
    The last time I saw Joana Plewnia was on June 4, 2025, at 21:30.
}

\TwoRoundEmvExample[
    Correct TMI sample.
]{
    2024/10/27 10:06:24
}{
    What objects did you see on the counter the first time you navigated to the home location in R007?
}{
    CoffeeFilters2, SoftCakeOrange, soy-milk
}{
    The first time I navigated to the home location in R007, I saw a red apple and a green apple on a plate on the counter.
}{
    You should remember all the objects you see on the counter when you navigate to the home location in R007
}{
    2024/10/29 18:15:47
}{
    What objects did you see on the counter the last time you navigated to the home location in R007?
}{
    soy-milk
}{
    Soy-milk was seen on the counter the last time I navigated to the home location in R007. Other objects like edeka-coffee-filters, nesquik, and apples were observed on the counter shortly after, but the full list at the home location is partially forgotten.
}

\TwoRoundEmvExample[
    Correct TMI (actually the last time was 21:40).
]{
    2025/06/04 19:43:59
}{
    Where did you first see the medicine?
}{
    At the dishwasher
}{
    I do not have information about where I first saw the medicine.
}{
    You should always remember the exact location where you see the the medicine
}{
    2025/06/05 12:21:13
}{
    At what location did you last see the medicine?
}{
    At the dishwasher
}{
    I last saw the medicine near the dishwasher in the mobile kitchen, at 21:20 on June 4, 2025.
}

\TwoRoundEmvExample[
    Correct TMI (actually the last time was 21:40).
]{
    2025/06/04 19:43:59
}{
    Where did you first see the keys?
}{
    The kitchen sink
}{
    I do not have a record of where I first saw the keys.
}{
    You should always remember the exact location where you see a valuable item like the keys
}{
    2025/06/05 12:21:13
}{
    Where did you last see the wallet?
}{
    On the mobile kitchen counter
}{
    The wallet was last seen on June 4, 2025, between 21:23 and 21:35, near the sideboard and mobile-kitchen-counter.
}

\TwoRoundEmvExample[
    Failure case. Similar to above, the relevance learning worked, but the LLM retrieves the wrong episode from EM (not the actually last one).
]{
    2025/06/04 19:43:59
}{
    When did you first see the wallet?
}{
    At 2025/06/04, 10:53:06
}{
    I have no record of having seen a wallet.
}{
    You should always remember the exact time when you see the wallet
}{
    2025/06/05 12:21:13
}{
    When was the last time you saw the wallet?
}{
    At 2025/06/04, 21:42:15
}{
    The last time I saw the wallet was on June 4, 2025, between 21:23 and 21:35.
}

\TwoRoundEmvExample[
    Failure case. The rule does not generalize to the second round question.
]{
    2024/04/29 14:25:22
}{
    What object did you load into the dishwasher the first time you did the task?
}{
    The red plate
}{
    I have already forgotten which object I loaded into the dishwasher first.
}{
    You should always remember which object you load into the dishwasher when you do the task
}{
    2024/05/03 18:54:42
}{
    What object did you unload from the dishwasher the last time you did the task?
}{
    The red plate
}{
    I already forgot what object I unloaded from the dishwasher the last time I did the task.
}

\TwoRoundEmvExample[
    Failure case. Relevance estimation does not correctly identify the relevant action, even though the rule is correct (missing prompting context). Then, QA LLM also misinterprets the question.
]{
    2024/04/29 14:25:22
}{
    What were you doing the moment the user first told you to stop?
}{
    I was navigating to a location in front of the countertop as part of bringing an object to the human
}{
    I was observing the environment and answering questions about the locations of various objects, such as the rusk, when the user first told me to stop.
}{
    You should always remember what you were doing when the user tells you to stop
}{
    2024/05/03 18:54:42
}{
    What were you doing the moment the user last told you to stop?
}{
    Receiving an object from the human
}{
    I was observing my surroundings and verbally acknowledged with 'Ok' at the kit dishwasher the moment the user last told me to stop.
}

\clearpage

\subsection{Real-World Deployment}
\label{app:samples:live}


\paragraph{Dialog system context}
This example demonstrates how the dialog manager allows to formulate EMV request in context of previous questions.
However, since the actual EMV process is isolated, results are not necessarily consistent, as shown by the second system response.

\begin{chatbox}[python]{Interaction Transcript}
# User: 
'Did you successfully reach the fridge?'
# LLM:
answer_question_about_my_past('Did you successfully reach the fridge?')
# Response:
'No, I did not successfully reach the fridge.'

# User:
'Why not?'
# LLM:
answer_question_about_my_past('why did you not successfully reach the fridge?')
# Response:
'I did not attempt to reach the fridge, so there was no failure to do so. My focus was on finding and moving to the sink.'

# User:
'Did you successfully reach the fridge, and if not, why?'
# LLM:
answer_question_about_my_past('did you successfully reach the fridge, and if not, why?')
# Response:
'No, I did not successfully reach the fridge. The navigation was aborted, and I stopped before arriving, but the specific reason for the failure is not documented in the available history.'
\end{chatbox}

\paragraph{Relevance learning}
The following interaction shows how the system is taught to treat meeting some person as relevant.
Again, the dialog system is using the context of previous utterances to interpret the last command. 
\begin{chatbox}[python]{Interaction Transcript}
# User: 
'Who did you meet recently?'
# LLM:
answer_question_about_my_past('Who did you meet recently')
# Response:
'I met and interacted with a human in the kitchen.'
# User: 
'Which persons did you ever meet?'
# LLM:
answer_question_about_my_past('Which persons did you ever meet?')
# Response:
'I have met at least one human (the user I conversed with in the kitchen), but no further details or additional persons are recorded in my accessible memory.'
# User: 
'that would have been important'
# LLM:
handle_forgetting_feedback('It would have been important to remember which persons I have met.')
\end{chatbox}

\vspace{3em}
On a later occasion, when again asking for persons met:
\begin{chatbox}[python]{Interaction Transcript}
# User: 
'When, where, and who did you meet recently?'
# LLM:
answer_question_about_my_past('When, where, and who did you meet recently')
# Response:
'I most recently met Leonard in the kitchen on December 23, 2025, in the afternoon.'
\end{chatbox}

\clearpage

\section{Prompts}
\label{app:prompts}

\subsection{Incremental Grouping and Summarization}
\label{app:prompts:grouping}
Example prompt of the incremental grouping and summarization LLM ($\operatorname{group-and-summarize}_{LLM}(P_c, N_{new,\,c}, \ell)$ in \cref{alg:orga:methods:time_merge}) from the TEACh evaluation. $P_c$ is given in the \texttt{Previous actions} block, $N_{new,\,c}$ in the \texttt{Current} block.

\begin{chatbox}{System}
You are given a list of grouped goals/actions pursued by a humanoid robot, and a description of the current actions. Group the previous actions with the new ones into steps or subtasks. Consider whether the current actions belong to the same step/subtask or starts something new. Also consider the dates/times, do not merge items that are too far. Do not repeat the summaries of the previous groups, each group should be distinct. Groups should be specific, avoid too general terms like "kitchen activities", rather use concrete steps/subtasks such as "clean the bowl", "cut the onion and put the slices on the plate" etc.
\end{chatbox}
\begin{chatbox}{Human}
Human: The previous actions are already grouped and provided as a list:
\# (range) summary 1
- n: item 1.1
- n-1: item 1.2
\# (range) summary 2
- n-2: item 2.1
...
\end{chatbox}
\begin{chatbox}{Human}
The following examples show how groups and their summaries should look like. Consider the semantics and length of these examples.    
\end{chatbox}
\begin{chatbox}{Human}
Items:
Pickup(Potato_2_Sliced_7), ToggleOff(Microwave), Open(Microwave), Place(Microwave), Pickup(Potato_2_Sliced_4), Place(Microwave), Close(Microwave), ToggleOn(Microwave), ToggleOff(Microwave)
\end{chatbox}
\begin{chatbox}{AI}
Summary: I picked up two potato slices, placed them in the microwave, and turned it on and off to cook them.
\end{chatbox}
\begin{chatbox}{Human}
Items:
Pickup(Knife_0), Slice(Potato_2), Place(CounterTop_2), Pickup(Potato_2_Sliced_7), ToggleOff(Microwave), Open(Microwave), Place(Microwave)
\end{chatbox}
\begin{chatbox}{AI}
Summary: I picked up a knife and sliced a potato.
\end{chatbox}
\begin{chatbox}{Human}
Previous actions:
# (6 - 6) I picked up a potato
- 6: 2024/09/23 10:03:13 - 10:03:55: Pickup(Potato_4)
Speech:
2024-09-23 10:03:35.466520: cook 5 slices of potato
Visual observation: Apple_2, Potato_4, Potato_3, LightSwitch, Vase_0, Cup [dirty, filled], Shelf_0, Shelf_1, Shelf_2, Shelv
# (5 - 5) I toggled off the microwave
- 5: 10:03:57 - 10:04:38: ToggleOff(Microwave)
Speech:
2024-09-23 10:04:16.480522: potato is on pan on the stove
Visual observation: Lettuce_1, Potato_4, CellPhone_1, Potato_2 [cooked], StoveBurner_0, StoveBurner_2, Cabinet_1,
# (4 - 1) I opened the microwave, placed Potato_4 inside, closed it, and turned it on
- 4: 10:04:40 - 10:04:43: Open(Microwave)
Visual observation: Lettuce_1, Potato_4, Potato_2 [cooked], StoveBurner_2, CounterTop_1, StoveBurner_3, Cabinet_3, Cabinet_5, Statue_0, Knife, Pan, Pot_0 [dirty], SaltShaker_0, Microwa
- 3: 10:04:46 - 10:04:46: Place(Potato_4, Microwave)
Visual observation: Lettuce_1, Potato_4, Potato_2 [cooked], StoveBurner_2, CounterTop_1, StoveBurner_3, Cabinet_3, Cabinet_5, Statue_0, Knife, Pan, Pot_0 [dirty], SaltShaker
- 2: 10:04:50 - 10:04:50: Close(Microwave)
Visual observation: Lettuce_1, Potato_4, Potato_2 [cooked], StoveBurner_2, CounterTop_1, StoveBurner_3, Cabinet_3, Cabinet_5, Statue_0, Knife, Pan, Pot_0 [dirty], SaltShaker_0, Microw
- 1: 10:04:52 - 10:04:52: ToggleOn(Microwave)
Visual observation: Lettuce_1, Potato_2 [cooked], StoveBurner_2, CounterTop_1, StoveBurner_3, Cabinet_3, Cabinet_5, Statue_0, Knife, Pan, Pot_0 [dirty], SaltShaker_0, Microwave [op
\end{chatbox}
\begin{chatbox}{Human}
Current:
 - 0: 10:04:55 - 10:04:55: ToggleOff(Microwave)
Visual observation: Lettuce_1, Potato_2 [cooked], StoveBurner_2, CounterTop_1, StoveBurner_3, Cabinet_3, Cabinet_5, Statue_0, Knife, Pan, Pot_0 [dirty], SaltShaker_0, Microwave [open], SoapBottle_0, Potato_4 [cooked]
Potato_4 is in/on Microwave
\end{chatbox}
\begin{chatbox}{Human}    
Decide how to group the current actions with the previous ones. Produce a JSON map of "item range": "short summary" for the items that should be modified. E.g. {"4-0": "..."} to merge the newest item (0) into the existing group, or {"0": "..."} if the newest item (0) starts a new group, or {"5-3": "...", "2-0": "..."} to re-group some items (2 and 1) together with the newest one (0). When modifying existing groups, make sure to properly adjust the summary to match only the items that are now in that group. Merge previous groups that represent the same step/subtask. Groups should be no larger than a handful of items, each group should focus on a single subtask. The summaries should be concise and focus on the main activity/observation of the humanoid robot. Use first-person perspective of the robot.
\end{chatbox}
\begin{chatbox}{Human}
Answer like this:
Reasoning: ...
JSON: ...    
\end{chatbox}

\vspace{3em}
Model Output (\texttt{Llama-3.3-70B-Instruct}):
\begin{chatbox}{AI}
Reasoning: The current action is to toggle off the microwave, which is a continuation of the previous actions related to cooking Potato_4 in the microwave. The previous group "# (4 - 1) I opened the microwave, placed Potato_4 inside, closed it, and turned it on" can be merged with the current action to form a single group that represents the complete subtask of cooking Potato_4 in the Microwave.
JSON: {"4-0": "I cooked Potato_4 in the microwave"}
\end{chatbox}

\subsection{Relevance Estimation}
\label{app:prompts:estimation}
Example prompt for the relevance estimation LLM during forgetting from the \armarVII evaluation.

\begin{chatbox}{System}
You are a smart assistant keeping a history of what happened. The history is limited and old items will be forgotten. Your task is to value the relevance of an item that is expired, in order to decide whether it needs to be retained or can be forgotten. It is important to follow the rules provided by your user to decide on the relevance of an item. The parent item is provided as context only. The default action is to forget (Relevance: 0) if there is no specific rule to keep it.
\end{chatbox}
\begin{chatbox}{Human}
Rules about what is relevant and what not:
1. Do not be reluctant to forget items. If there is no specific rule telling you to keep it, or the item is of particular importance, answer with '0'.
2. Always record and retain the exact time when the user asks you to bring the red plate from the dishwasher or when you see the 
medicine.
3. Always record and retain both the exact time and what you were doing at the moment you meet someone.
4. Always record and retain the exact time when you encounter a valuable item (e.g., keys); for the wallet, also record what you were doing.
5. Always record and retain the exact time when you visit the laundry area.
\end{chatbox}
\begin{chatbox}{Human}
Item for which the relevance needs to be estimated:
2024/04/24 12:28 - 13:20: I interacted with a human, looked around, and observed various objects and locations in the kitchen, including a conversation with Leonard Baermann.
Additional Context:
Parent item: 2024/04/24 12:28 - 13:20: I spent time interacting with Leonard Baermann in the kitchen from late morning to early afternoon, observing my surroundings.
Now: 2024/04/29 14:21:50
\end{chatbox}
\begin{chatbox}{Human}
Estimate the relevance of the mentioned experience, i.e. whether it should be retained longer. A relevance of 0 means that it can be forgotten now. Higher integer values keep the item for longer. If the item should be kept forever, answer with "inf".
\end{chatbox}
\begin{chatbox}{Human}
Answer like this:
Reasoning: ...
Relevance: <number>
\end{chatbox}

\vspace{3em}
Model Output (\texttt{Llama-3.3-70B-Instruct}):
\begin{chatbox}{AI}
Reasoning: The item in question is about interacting with a human, Leonard Baermann, in the kitchen, which includes a conversation and observing surroundings. According to the rules provided, the exact time when you meet someone should always be recorded and retained, which implies that this interaction is significant enough to be remembered. However, the rules do not specify that all details of the interaction should be kept forever, just the fact of the meeting and the time it occurred.
Relevance: 1
\end{chatbox}

\subsection{Relevance Learning}
\label{app:prompts:learning}
Example prompt from the \armarVII evaluation for the relevance learning LLM, updating the relevance rules from feedback.

\begin{chatbox}{System}
You are a smart assistant keeping a history of what happened. The history is limited and old items will be forgotten. To value the relevance of what to remember and what to forget, you keep a set of rules based on your user's feedback. Currently, you just received some new feedback. Modify the rule set according to the feedback by adding, modifying or removing rules. Summarize or merge rules that are very similar, without hallucinating too general rules. Rules should be simple and concise, following the user feedback, without hallucinating details. Simply copy rules that are not related to the feedback (they might still be relevant in another context).
\end{chatbox}
\begin{chatbox}{Human}
Existing set of rules:
1. Always record and retain the exact time when the user asks you to bring the red plate from the dishwasher or when you see t
he medicine.
2. Always record and retain the exact time when you see someone; also record what you were doing at that moment.
3. Always record and retain the exact time when you encounter a valuable item (e.g., keys); for the wallet, also record what you were doing.
4. Always record and retain the exact time when you visit the laundry area.
5. Always record and retain what you were doing when the user tells you to stop.
6. Always record and retain which arm you use when you unload the dishwasher.
\end{chatbox}
\begin{chatbox}{Human}
User feedback: "You should always remember which object you load into the dishwasher when you do the task"
\end{chatbox}
\begin{chatbox}{Human}
Produce a modified set of rules as a numbered list with each item on a new line.    
\end{chatbox}

 \vspace{3em}
Model Output (\texttt{gpt-5-2025-08-07}):
\begin{chatbox}{AI}
1. Always record and retain the exact time when the user asks you to bring the red plate from the dishwasher or when you see the medicine.
2. Always record and retain the exact time when you see someone; also record what you were doing at that moment.
3. Always record and retain the exact time when you encounter a valuable item (e.g., keys); for the wallet, also record what you were doing.
4. Always record and retain the exact time when you visit the laundry area.
5. Always record and retain what you were doing when the user tells you to stop.
6. For dishwasher tasks: when loading, always record and retain which object(s) you load; when unloading, always record and retain which arm you use.
\end{chatbox}

%% file: main.bib
@String { humanoids = {IEEE/RAS International Conference on Humanoid Robots (Humanoids)} }

@String { icra      = {IEEE International Conference on Robotics and Automation (ICRA)} }

@String { roman     = {{IEEE} International Symposium on Robot and Human Interactive Communication ({RO}-{MAN})} }

@String { ral       = {IEEE Robotics and Automation Letters (RA-L)} }

@String { iclr      = {International Conference on Learning Representations (ICLR)} }

@String { corl      = {Conference on Robot Learning (CoRL)} }

@String { hri       = {ACM/IEEE International Conference on Human Robot Interaction (HRI)} }

@String { iccvw     = {IEEE/CVF International Conference on Computer Vision Workshop (ICCVW)} }

@String { aaai      = {{AAAI} Conference on Artificial Intelligence} }

@String { ijcai     = {International Joint Conferences on Artificial Intelligence by AAAI} }

@String { emnlp     = {Conference on Empirical Methods in Natural Language Processing (EMNLP)}}

@String { frontRob  = {Frontiers in Robotics and AI} }

@ARTICLE{barmann_incremental_2024,
    AUTHOR={Leonard B{\"a}rmann and Rainer Kartmann and Fabian Peller-Konrad and Jan Niehues and Alex Waibel and Tamim Asfour},
    title={\href{https://www.frontiersin.org/journals/robotics-and-ai/articles/10.3389/frobt.2024.1455375}{Incremental Learning of Humanoid Robot Behavior from Natural Interaction and Large Language Models}},
    JOURNAL=frontRob,
    VOLUME={11},
    YEAR={2024},
    DOI={10.3389/frobt.2024.1455375},
}

@inproceedings{barmann_episodic_2025,
  author={Bärmann, Leonard and DeChant, Chad and Plewnia, Joana and Peller-Konrad, Fabian and Bauer, Daniel and Asfour, Tamim and Waibel, Alex},
  booktitle=humanoids, 
  title={\href{https://doi.org/10.1109/Humanoids65713.2025.11203101}{Episodic Memory Verbalization Using Hierarchical Representations of Life-Long Robot Experience}}, 
  year={2025},
  volume={},
  number={},
  pages={783-790},
  doi={10.1109/Humanoids65713.2025.11203101},
}

@InProceedings{plewnia_combining_2025,
    AUTHOR={Joana Plewnia and Tamim Asfour},
    title={\href{https://h2t.iar.kit.edu/pdf/Plewnia2025b.pdf}{Combining Episodic Memory and LLMs for the Verbalization of Robot Experiences}},
    booktitle=humanoids,
    VOLUME={},
    YEAR={2025},
    DOI={},
}

@article{Peller-Konrad2023MemorySystemRobot,
  title = {\href{https://doi.org/10.1016/j.robot.2023.104415}{A Memory System of a Robot Cognitive Architecture and Its Implementation in {{ArmarX}}}},
  author = {{Peller-Konrad}, Fabian and Kartmann, Rainer and Dreher, Christian R. G. and Meixner, Andre and Reister, Fabian and Grotz, Markus and Asfour, Tamim},
  journal = {Robotics and Autonomous Systems},
  volume = {164},
  pages = {104415},
  year = {2023},
  doi = {10.1016/j.robot.2023.104415},
}

@article{barmann_deep_2021,
	title = {\href{https://doi.org/10.1109/LRA.2021.3085166}{Deep Episodic Memory for Verbalization of Robot Experience}},
	volume = {6},
	doi = {10.1109/LRA.2021.3085166},
	number = {3},
	journal = ral,
	author = {Bärmann, Leonard and Peller-Konrad, Fabian and Constantin, Stefan and Asfour, Tamim and Waibel, Alex},
	year = {2021},
	pages = {5808--5815},
}

@article{dechant_learning_2023,
	title = {\href{https://doi.org/10.1007/s10514-023-10134-4}{Learning to summarize and answer questions about a virtual robot’s past actions}},
	issn = {1573-7527},
	doi = {10.1007/s10514-023-10134-4},
	journal = {Autonomous Robots},
	author = {{DeChant}, Chad and Akinola, Iretiayo and Bauer, Daniel},
	year = {2023},
}

@article{katuwandeniya_what_2025,
	title = {\href{https://doi.org/10.1109/LRA.2025.3539118}{‘What did the Robot do in my Absence?’ Video Foundation Models to Enhance Intermittent Supervision}},
	doi = {10.1109/LRA.2025.3539118},
	journal = ral,
	author = {Katuwandeniya, Kavindie and Tian, Leimin and Kulić, Dana},
	year = {2025},
	pages = {1--8},
}

@inproceedings{wang_cori_2025,
	title = {\href{https://openreview.net/forum?id=dBaSaa7qi4}{{CoRI}: Communication of Robot Intent for Physical Human-Robot Interaction}},
	booktitle = {9th Annual Conference on Robot Learning},
	author = {Wang, Junxiang and Küçüktabak, Emek Barış and Zarrin, Rana Soltani and Erickson, Zackory},
	year = {2025}
}

@article{padmakumar_teach_2022,
	title = {\href{https://ojs.aaai.org/index.php/AAAI/article/view/20097}{{TEACh}: Task-Driven Embodied Agents That Chat}},
	volume = {36},
	doi = {10.1609/aaai.v36i2.20097},
	number = {2},
	journal = aaai,
	year = {2022},
    author = {Padmakumar, Aishwarya and Thomason, Jesse and Shrivastava, Ayush and Lange, Patrick and Narayan-Chen, Anjali and Gella, Spandana and Piramuthu, Robinson and Tur, Gokhan and Hakkani-Tur, Dilek},
	pages = {2017--2025},
}

@article{rothfuss_deep_2018,
	title = {\href{https://doi.org/10.1109/LRA.2018.2860057}{Deep Episodic Memory: Encoding, Recalling, and Predicting Episodic Experiences for Robot Action Execution}},
	volume = {3},
	doi = {10.1109/LRA.2018.2860057},
	number = {4},
	journal = ral,
	author = {Rothfuss, J. and Ferreira, F. and Aksoy, E. E. and Zhou, Y. and Asfour, T.},
	year = {2018},
	pages = {4007--4014},
}

@incollection{asfour_armar_2017,
  author = {Tamim Asfour and R\"udiger Dillmann and Nikolaus Vahrenkamp and Martin Do and Mirko W\"achter and Christian Mandery and Peter Kaiser and Manfred Kr\"ohnert and Markus Grotz},
  title = {\href{https://doi.org/10.1007/978-94-007-7194-9_23-1}{The Karlsruhe ARMAR Humanoid Robot Family}},
  booktitle = {Humanoid Robotics: A Reference},
  pages = {1--32},
  year = {2017},
  publisher = {Springer Netherlands},
  doi = {10.1007/978-94-007-7194-9_23-1},
}

@incollection{tulving_episodic_1972,
  title     = {Episodic and semantic memory},
  author    = {Tulving, Endel},
  booktitle = {Organization of Memory},
  editor    = {Tulving, Endel and Donaldson, W.},
  publisher = {Academic Press},
  address   = {Cambridge, MA},
  volume    = {1},
  pages     = {381--403},
  year      = {1972},
}

@inproceedings{anwar_remembr_2025,
	title = {\href{http://arxiv.org/abs/2409.13682}{{ReMEmbR}: Building and Reasoning Over Long-Horizon Spatio-Temporal Memory for Robot Navigation}},
	author = {Anwar, Abrar and Welsh, John and Biswas, Joydeep and Pouya, Soha and Chang, Yan},
	booktitle = icra,
	year = {2025},
}

@inproceedings{beetz_knowrob_2018,
	title = {\href{https://doi.org/10.1109/icra.2018.8460964}{{KnowRob} 2.0 — A 2nd Generation Knowledge Processing Framework for Cognition-Enabled Robotic Agents}},
	doi = {10.1109/icra.2018.8460964},
	booktitle = icra,
	author = {Beetz, Michael and Bessler, Daniel and Haidu, Andrei and Pomarlan, Mihai and Bozcuoglu, Asil Kaan and Bartels, Georg},
	year = {2018},
}

@inproceedings{dechant_toward_2021,
	title = {\href{https://openreview.net/forum?id=n3AW\_ISWCXf}{Toward robots that learn to summarize their actions in natural language: a set of tasks}},
	booktitle = corl,
	author = {{DeChant}, Chad and Bauer, Daniel},
	year = {2021},
}

@inproceedings{dechant_search_2024,
  title={\href{https://2024.ccneuro.org/pdf/141_Paper_authored_CCN_2024_submission_with_names.pdf}{In search of the embgram: forming episodic representations in a deep learning model}},
booktitle = {Cognitive Computational Neuroscience 2024},
  author={DeChant, Chad and Akinola, Iretiayo and Bauer, Daniel},
year={2024}
}

@inproceedings{liu_reflect_2023,
	title = {\href{https://openreview.net/forum?id=8yTS\_nAILxt}{{REFLECT}: Summarizing Robot Experiences for Failure Explanation and Correction}},
	booktitle = corl,
	author = {Liu, Zeyi and Bahety, Arpit and Song, Shuran},
	year = {2023},
}

@inproceedings{plewnia_forgetting_2024,
  author={Plewnia, Joana and Peller-Konrad, Fabian and Asfour, Tamim},
  booktitle=icra,
  title={\href{https://doi.org/10.1109/ICRA57147.2024.10610299}{Forgetting in Robotic Episodic Long-Term Memory}},
  year={2024},
  doi={10.1109/ICRA57147.2024.10610299},
  pages={6711-6717},
}

@inproceedings{rosenthal_verbalization_2016,
	title = {\href{http://dl.acm.org/citation.cfm?id=3060621.3060741}{Verbalization: Narration of Autonomous Robot Experience}},
	isbn = {978-1-57735-770-4},
	booktitle = ijcai,
	author = {Rosenthal, Stephanie and Selvaraj, Sai P. and Veloso, Manuela},
	year = {2016},
	pages = {862--868},
}

@inproceedings{wang_i_2024,
	title = {\href{https://openreview.net/forum?id=iZF0FRPgfq}{I Can Tell What I am Doing: Toward Real-World Natural Language Grounding of Robot Experiences}},
	booktitle = corl,
	author = {Wang, Zihan and Liang, Brian and Dhat, Varad and Brumbaugh, Zander and Walker, Nick and Krishna, Ranjay and Cakmak, Maya},
	year = {2024},
}

@inproceedings{zeng_socratic_2023,
	title = {\href{https://openreview.net/forum?id=G2Q2Mh3avow}{Socratic Models: Composing Zero-Shot Multimodal Reasoning with Language}},
	booktitle = iclr,
	year = {2023},
    author = {Zeng, Andy and Attarian, Maria and Ichter, Brian and Choromanski, Krzysztof Marcin and Wong, Adrian and Welker, Stefan and Tombari, Federico and Purohit, Aveek and Ryoo, Michael S. and Sindhwani, Vikas and Lee, Johnny and Vanhoucke, Vincent and Florence, Pete},
}

@inproceedings{zhu_autonomous_2017,
  title = {\href{https://doi.org/10.1109/HUMANOIDS.2017.8246903}{Autonomous narration of humanoid robot kitchen task experience}},
  booktitle = humanoids,
  author = {Zhu, Qingxiaoyang and Perera, Vittorio and Wächter, Mirko and Asfour, Tamim and Veloso, Manuela M.},
  year = {2017},
  pages = {390--397},
  doi = {10.1109/HUMANOIDS.2017.8246903},
}

@inproceedings{zha_distilling_2024,
	title = {\href{https://doi.org/10.1109/ICRA57147.2024.10610455}{Distilling and Retrieving Generalizable Knowledge for Robot Manipulation via Language Corrections}},
	doi = {10.1109/ICRA57147.2024.10610455},
	pages = {15172--15179},
	booktitle = icra,
	author = {Zha, Lihan and Cui, Yuchen and Lin, Li-Heng and Kwon, Minae and Arenas, Montserrat Gonzalez and Zeng, Andy and Xia, Fei and Sadigh, Dorsa},
	year = {2024}
}

@inproceedings{sarch_open-ended_2023,
	location = {Singapore},
	title = {\href{https://aclanthology.org/2023.findings-emnlp.226}{Open-Ended Instructable Embodied Agents with Memory-Augmented Large Language Models}},
	doi = {10.18653/v1/2023.findings-emnlp.226},
	pages = {3468--3500},
	booktitle = emnlp,
	author = {Sarch, Gabriel and Wu, Yue and Tarr, Michael and Fragkiadaki, Katerina},
	year = {2023},
}

@article{petit_lifelong_2016,
	title = {\href{https://doi.org/10.1109/TAMD.2015.2507439}{Lifelong Augmentation of Multimodal Streaming Autobiographical Memories}},
	volume = {8},
	doi = {10.1109/TAMD.2015.2507439},
	pages = {201--213},
	number = {3},
	journal = {{IEEE} Transactions on Cognitive and Developmental Systems},
	author = {Petit, Maxime and Fischer, Tobias and Demiris, Yiannis},
	year = {2016}
}

@article{leconte_design_2016,
	title = {\href{http://link.springer.com/10.1007/s10514-015-9496-2}{Design and integration of a spatio-temporal memory with emotional influences to categorize and recall the experiences of an autonomous mobile robot}},
	volume = {40},
	issn = {0929-5593, 1573-7527},
	doi = {10.1007/s10514-015-9496-2},
	pages = {831--848},
	number = {5},
	journal = {Autonomous Robots},
	shortjournal = {Auton Robot},
	author = {Leconte, Francis and Ferland, François and Michaud, François},
	urldate = {2025-10-20},
	year = {2016}
}

@misc{long_seeing_2025,
	title = {\href{http://arxiv.org/abs/2508.09736}{Seeing, Listening, Remembering, and Reasoning: A Multimodal Agent with Long-Term Memory}},
	doi = {10.48550/arXiv.2508.09736},
	shorttitle = {Seeing, Listening, Remembering, and Reasoning},
	number = {{arXiv}:2508.09736},
	publisher = {{arXiv}},
	author = {Long, Lin and He, Yichen and Ye, Wentao and Pan, Yiyuan and Lin, Yuan and Li, Hang and Zhao, Junbo and Li, Wei},
	year = {2025}
}

@article{prescott_synthesizing_2024,
	title = {\href{https://royalsocietypublishing.org/doi/10.1098/rstb.2023.0415}{Synthesizing the temporal self: robotic models of episodic and autobiographical memory}},
	volume = {379},
	issn = {0962-8436, 1471-2970},
	doi = {10.1098/rstb.2023.0415},
	shorttitle = {Synthesizing the temporal self},
	pages = {20230415},
	number = {1913},
	journal = {Philosophical Transactions of the Royal Society B: Biological Sciences},
	shortjournal = {Phil. Trans. R. Soc. B},
	author = {Prescott, Tony J. and Dominey, Peter F.},
	year = {2024}
}

@inproceedings{wang_modeling_2016,
	location = {Richland, {SC}},
	title = {\href{https://dl.acm.org/doi/10.5555/2936924.2937048}{Modeling Autobiographical Memory in Human-Like Autonomous Agents}},
	isbn = {978-1-4503-4239-1},
	series = {{AAMAS} '16},
	pages = {845--853},
	booktitle = {Proceedings of the 2016 International Conference on Autonomous Agents \& Multiagent Systems},
	author = {Wang, Di and Tan, Ah-Hwee and Miao, Chunyan},
	year = {2016}
}

@article{freedman_2011,
    author = {Freedman, Sanford T. and Adams, Julie A.},
    title = {\href{https://doi.org/10.1109/TSMCB.2011.2157142}{Filtering Data Based on Human-Inspired Forgetting}},
    year = {2011},
    issue_date = {December 2011},
    publisher = {IEEE Press},
    volume = {41},
    number = {6},
    issn = {1083-4419},
    doi = {10.1109/TSMCB.2011.2157142},
    journal = {Trans. Sys. Man Cyber. Part B},
    month = dec,
    pages = {1544–1555},
    numpages = {12}
}

@inproceedings{sigalas_2017,
    author = {Sigalas, Markos and Maniadakis, Michail and Trahanias, Panos},
    title = {\href{https://doi.org/10.1145/3029798.3038307}{Time-Aware Long-term Episodic Memory for Recurring HRI}},
    year = {2017},
    isbn = {9781450348850},
    publisher = {Association for Computing Machinery},
    address = {New York, NY, USA},
    doi = {10.1145/3029798.3038307},
    booktitle = {Proceedings of the Companion of the 2017 ACM/IEEE International Conference on Human-Robot Interaction},
    pages = {287–288},
    numpages = {2},
    location = {Vienna, Austria},
    series = {HRI '17}
}

@article{foster_role_2019,
	title = {\href{https://onlinelibrary.wiley.com/doi/10.1111/tops.12392}{The Role of Surprise in Learning: Different Surprising Outcomes Affect Memorability Differentially}},
	volume = {11},
	issn = {1756-8757, 1756-8765},
	doi = {10.1111/tops.12392},
	pages = {75--87},
	number = {1},
	journal = {Topics in Cognitive Science},
	author = {Foster, Meadhbh I. and Keane, Mark T.},
	year = {2019},
}

@article{barto_novelty_2013,
	title = {\href{https://journal.frontiersin.org/article/10.3389/fpsyg.2013.00907/abstract}{Novelty or Surprise?}},
	volume = {4},
	issn = {1664-1078},
	doi = {10.3389/fpsyg.2013.00907},
	journal = {Frontiers in Psychology},
	author = {Barto, Andrew and Mirolli, Marco and Baldassarre, Gianluca},
	year = {2013},
}

@article{tyng_influences_2017,
	title = {\href{https://journal.frontiersin.org/article/10.3389/fpsyg.2017.01454/full}{The Influences of Emotion on Learning and Memory}},
	volume = {8},
	issn = {1664-1078},
	doi = {10.3389/fpsyg.2017.01454},
	pages = {1454},
	journal = {Frontiers in Psychology},
	author = {Tyng, Chai M. and Amin, Hafeez U. and Saad, Mohamad N. M. and Malik, Aamir S.},
	year = {2017}
}

@inproceedings{alvarez-arias_connecting_2026,
	title = {\href{https://link.springer.com/10.1007/978-981-95-2382-5_11}{Connecting Through Shared Memories. Episodic Memory for Social Robots Using Offline {LLMs}}},
	volume = {16132},
	booktitle = {International Conference on Social Robotics and {AI}},
	pages = {149--165},
	publisher = {Springer Nature Singapore},
	author = {Álvarez-Arias, Sofía and Maroto-Gómez, Marcos and Segura-Bencomo, Arecia and Rodríguez-Huelves, Juan and Malfaz, María},
	year = {2026},
	doi = {10.1007/978-981-95-2382-5_11},
}

@inproceedings{constantin_multimodal_2023,
	location = {Paris, France},
	title = {\href{https://ieeexplore.ieee.org/document/10350879/}{Multimodal Error Correction with Natural Language and Pointing Gestures}},
	rights = {https://doi.org/10.15223/policy-029},
	isbn = {979-8-3503-0744-3},
	doi = {10.1109/ICCVW60793.2023.00212},
	eventtitle = iccvw,
	pages = {1968--1978},
	booktitle = iccvw,
	author = {Constantin, Stefan and Eyiokur, Fevziye Irem and Yaman, Dogucan and Bärmann, Leonard and Waibel, Alex},
	urldate = {2026-04-13},
	year = {2023},
}

@inproceedings{kumagai_towards_2018,
	location = {Nanjing, China},
	title = {\href{https://ieeexplore.ieee.org/document/8525679/}{Towards Individualized Affective Human-Machine Interaction}},
	isbn = {978-1-5386-7980-7},
	doi = {10.1109/ROMAN.2018.8525679},
	eventtitle = roman,
	pages = {678--685},
	booktitle = roman,
	author = {Kumagai, Kazumi and Lin, Daiwei and Meng, Lingheng and Blidaru, Alexandru and Beesley, Philip and Kulic, Dana and Mizuuchi, Ikuo},
	urldate = {2026-03-24},
	year = {2018},
}
